\documentclass[11pt]{article}

\usepackage[square,numbers]{natbib}
\usepackage{url}

\usepackage{fullpage}
\usepackage{times}
\usepackage[parfill]{parskip}
\usepackage[utf8]{inputenc}
\usepackage[T1]{fontenc}
\usepackage{microtype}

\usepackage{chemformula} 

\usepackage{graphicx}
\usepackage{adjustbox}
\usepackage{subcaption}
\usepackage{caption}
\usepackage{multirow}
\usepackage{threeparttable}
\usepackage{tablefootnote}
\usepackage{placeins}
\usepackage{float}
\usepackage{booktabs}

\usepackage{amsmath}
\usepackage{amssymb}
\usepackage{mathtools}
\usepackage{amsthm}
\usepackage{bm}
\usepackage{eucal}
\usepackage{cancel}
\usepackage{xspace}

\usepackage{hyperref}
\usepackage[capitalize,noabbrev]{cleveref}

\usepackage{algorithm}
\usepackage{algpseudocode}
\algdef{SE}[SUBALG]{Indent}{EndIndent}{}{\algorithmicend\ }%
\algtext*{Indent}
\algtext*{EndIndent}

\usepackage{enumitem}
\usepackage{xcolor,colortbl}

\theoremstyle{plain}

\theoremstyle{definition}

\theoremstyle{remark}

\newcommand{\model}{\textnormal{\textbf{BADGER} }}
\newcommand{\modelnobf}{\textnormal{BADGER }}
\newcommand{\modelnobfns}{\textnormal{BADGER}}
\newcommand{\modelns}{\textnormal{\textbf{BADGER}}}

\newcommand{\vfuncns}{\textnormal{ADV energy function}}
\newcommand\sref{\S\ref}

\newcommand\eref{Eq.~\ref}
\newcommand\fref{Fig.~\ref}
\newcommand\tref{Tab.~\ref}
\newcommand\supsref{supplementary \S}
\newcommand\mref{ main paper~\S}
\newcommand\mtref{ main paper Tab.}

\title{\textbf{General Binding Affinity Guidance for Diffusion Models in Structure-Based Drug Design}}

\author{%
  Yue Jian\thanks{Co-first author.} \\ 
  University of California, Berkeley \\
  \texttt{yue\_jian@berkeley.edu}
  \and Curtis Wu\footnotemark[1] \\
  University of California, Berkeley \\
  \texttt{curtis\_wu@berkeley.edu}
  \and Danny Reidenbach \\
  NVIDIA \\
  \texttt{dreidenbach@nvidia.com}
  \and Aditi S. Krishnapriyan \\
  University of California, Berkeley \\
  \texttt{aditik1@berkeley.edu}
}
\date{}

\begin{document}
\maketitle

\begin{abstract}
Structure-based drug design (SBDD) aims to generate ligands that bind strongly and specifically to target protein pockets. Recent diffusion models have advanced SBDD by capturing the distributions of atomic positions and types, yet they often underemphasize binding affinity control during generation. To address this limitation, we introduce \textbf{\modelns}, a general \textbf{binding-affinity guidance framework for diffusion models in SBDD}. \model incorporates binding affinity awareness through two complementary strategies: (1) \textit{classifier guidance}, which applies gradient-based affinity signals during sampling in a plug-and-play fashion, and (2) \textit{classifier-free guidance}, which integrates affinity conditioning directly into diffusion model training. Together, these approaches enable controllable ligand generation guided by binding affinity. \model can be added to any diffusion model and achieves up to a \textbf{60\% improvement in ligand--protein binding affinity} of sampled molecules over prior methods. Furthermore, we extend the framework to \textbf{multi-constraint diffusion guidance}, jointly optimizing for binding affinity, drug-likeness (QED), and synthetic accessibility (SA) to design realistic and synthesizable drug candidates.

\end{abstract}
\section{Introduction}
Structure-based drug design (SBDD) is a fundamental task in drug discovery, aimed at designing ligand molecules that have a high binding affinity to the receptor protein pocket~\cite{anderson2003process,blundell1996structure}. SBDD directly utilizes the three-dimensional structures of target proteins, enabling the design of molecules that can specifically interact with and influence the activity of these proteins, thus increasing the specificity and efficacy of potential drugs. A major challenge in SBDD is to design ligand molecules with desired properties, which includes properties like the binding affinity, as well as other physical and chemical properties.
Conventional SBDD achieves this by conducting a ``filtering'' and ``optimization'' style workflow. During the filtering stage, a protein pocket is pre-selected and fixed, and a large database of ligand molecules is searched to find promising candidates with molecular docking. In the subsequent optimization stage, these candidate ligands are refined to improve desired properties using quantum mechanics or molecular mechanics (QM/MM) methods, combined with expert intuition and experience~\cite{alhossary2015fast,trott2010autodock,halgren2004glide}. However, this type of workflow faces several challenges. First, high-throughput experimental techniques or computational methods are both time-consuming and computationally demanding. Second, the search space for potential drug molecules is confined to the chemical database used in SBDD, limiting the diversity of candidates. Third, the optimization of candidate molecules post-identification is often influenced by human experience, which can introduce biases. These issues
highlight the need for novel methods in SBDD to address these limitations effectively.

Recent advances in machine learning, particularly in generative modeling, have provided a potentially computationally efficient alternative to the traditional SBDD approach. These developments can help overcome the limitations associated with the extensive ligand screening databases traditionally used in SBDD~\cite{guan20233d,xu2022geodiff,raja2025actionminimization,hoogeboom2022equivariant,reidenbach2024coarsenconf,reidenbach2024evosbdd,gao2020}. Among the various types of generative models used for SBDD, diffusion models have shown some success in generating ligands that have high binding affinity to their target protein pockets~\cite{dhariwal2021diffusion,guan20233d,guan2023decompdiff,schneuing2022structure}. In SBDD, diffusion model approaches model the continuous distribution of atom coordinates and discrete distribution of element types for both the ligand and protein. \citet{schneuing2022structure} and \citet{guan20233d} first introduced diffusion models for SBDD by conditioning both the training and sampling stages of ligand molecule design on the protein pocket. Building on this idea, \citet{guan2023decompdiff} developed a fragment-based strategy to improve ligand validity and binding affinity. Their approach decomposes ligands into fragments and initializes fragment positions using pre-designed priors before sampling. However, the effectiveness of this method strongly depends on the type and quality of the priors, which are tailored to specific pocket and ligand families, thereby limiting its generalizability to new systems. To further improve pocket–ligand binding affinity, \citet{zhou2023decompopt} proposed a filtering-based approach that incorporates physics-based predictors, such as AutoDock Vina’s scoring function (\vfuncns), during sampling. This method ranks and selects top candidates based on predicted binding affinity, but achieving substantial gains requires generating a large pool of ligands for filtering, which increases computational demands. Thus, developing a general and efficient strategy for integrating binding affinity into diffusion models for SBDD remains an open challenge.

A key advantage of diffusion models is their ability to perform conditional sampling, where the model can be conditioned on specific properties during the training or sampling stages. This capability has been extensively validated in fields like image generation~\cite{dhariwal2021diffusion,song2020score}. In the context of SBDD, applying conditional sampling strategies holds promise. By conditioning the model on desired ligand properties, diffusion models can facilitate the one-step design of ligands with optimal characteristics for a given protein pocket, offering a pocket agnostic, streamlined and highly targeted approach to drug design.

There are two commonly used conditional sampling strategies in diffusion models: classifier guidance and classifier-free guidance, each applied at different stages of the diffusion process. Classifier guidance achieves conditional sampling by utilizing the gradients of separately trained classifiers to iteratively steer the sampled data toward regions with desired properties during the sampling stage. This approach is a plug-and-play post-training method that does not require retraining the diffusion model. In contrast, classifier-free guidance operates during the training phase. It involves training a single diffusion model that jointly models both conditional and unconditional distributions by randomly dropping the condition label during training. While this method requires modifications to the diffusion model’s training process, it eliminates the need for a separate classifier to provide guidance. Meanwhile, this method does not explicitly introduce additional computational cost during the sampling stage compared to unconditional diffusion sampling.
 In most image generation tasks, the condition for these methods is typically a class label represented by a multinomially distributed vector. However, in drug design, many relevant ligand properties, such as binding affinity, are continuous scalars distributed over specific intervals with physical meaning. Binding affinity is a critical measure of how effectively a ligand interacts with a protein pocket. In practice, binding affinity is often approximated using scoring functions like AutoDock Vina’s energy function (denoted as \vfuncns), which estimates binding free energy based on atomic interactions~\cite{trott2010autodock}.

We introduce \textbf{\modelnobfns}, a general guidance framework for designing ligands with desired properties in diffusion models for structure-based drug design (SBDD), with a particular emphasis on improving protein–ligand binding affinity. The core idea of \modelnobf is to model and sample 3D ligand structures from distributions conditioned on specific molecular properties. \modelnobf comprises two complementary variants: \textit{classifier guidance} and \textit{classifier-free guidance}. The classifier-guided variant leverages gradients from a separately trained property predictor to steer the diffusion sampling process toward regions of higher binding affinity. This plug-and-play approach requires no modification to diffusion model training and can be readily applied to existing SBDD frameworks. In contrast, the classifier-free variant integrates property conditioning directly into the diffusion model during training, enabling both conditional and unconditional generation within a single model. By interpolating between these two modes, classifier-free guidance produces ligands that better align with target properties and exhibit improved binding performance. 

We evaluate \modelnobf on the CrossDocked2020~\cite{francoeur2020three} and PDBBind~\cite{wang2005pdbbind} datasets, demonstrating improvements in ligand binding affinity compared to baseline diffusion models. Beyond single-objective optimization, we extend \modelnobf to \textbf{multi-property diffusion guidance}, jointly optimizing for binding affinity, quantitative estimate of drug-likeness (QED)~\cite{bickerton2012quantifying}, and synthetic accessibility (SA)~\cite{ertl2009estimation}. We further show that incorporating affinity guidance improves binding selectivity, as measured by the binding specificity score~\cite{gao2024rethinking}. The code for this work will be publicly available at \url{https://github.com/ASK-Berkeley/BADGER-SBDD}.

\subsection{Problem definition}\label{pdsec}
\paragraph{Structure-based Drug Design.}
Consider a protein pocket with $N_p$ atoms, where each atom is described by $N_f$ feature dimensions. We represent this as a matrix $P =[\boldsymbol{x_{p}},\boldsymbol{v_{p}}]$, where $\boldsymbol{x_p} \in \mathbb{R}^{N_p \times 3}$ represents the Cartesian coordinates of the atoms, and $\boldsymbol{v_p} \in \mathbb{R}^{N_p \times N_f}$ represents the atom features for atoms that form the protein pocket. We define the operation $[\cdot,\cdot]$ to be concatenation. Let a ligand molecule with $N_m$ atoms, each also described by $N_f$ feature dimensions, be represented as matrix $M = [\boldsymbol{x}, \boldsymbol{v}]$, where $\boldsymbol{x} \in \mathbb{R}^{N_m \times 3}$ and $\boldsymbol{v} \in \mathbb{R}^{N_m \times N_f}$.

The binding affinity between the protein pocket $P$ and the ligand molecule $M$ is denoted by $\Delta G(P,M)$. In the context of SBDD, the goal is to generate ligand $M$, given a protein pocket $P$, such that $\Delta G(P,M) < 0$. A more negative value of $\Delta G(P,M)$ indicates a stronger and more favorable binding interaction between the ligand and the protein, which is a desirable property in drug discovery.

\paragraph{Problem of Interest.} Building on this background, we are interested in improving the binding affinity $\Delta G(P,M)$, specifically by generating ligands $M$ that achieve a lower $\Delta G(P,M)$ using diffusion-based SBDD methods. In our approach, we use diffusion models tailored for SBDD. Our goal is to develop a guidance strategy for the diffusion model that enables the generation of molecules with higher binding affinity when the guidance is employed, ideally achieving $\Delta G_{guided} < \Delta G_{unguided}$. Furthermore, we want to generalize this to other properties like QED and SA, seeking to improve $\Delta G$, QED and SA for sampled ligand $M$ within one sampling path simultaneously.

\subsection{Diffusion Models for Structure-based Drug Design}\label{dfsbdd}
Recent advancements in generative modeling have been effectively applied to the SBDD task~\cite{peng2022pocket2mol,liu2022generating,luo20213d}. The development of denoising diffusion probabilistic models~\cite{song2019generative,song2020score,ho2020denoising,dhariwal2021diffusion} has led to approaches in SBDD using diffusion models~\cite{guan20233d,guan2023decompdiff,zhou2023decompopt}. 

In the current literature of diffusion models for SBDD, both protein pockets and ligands are modeled as point clouds. In the sampling stage, protein pockets are treated as the fixed ground truth across all time steps, while ligands start as Gaussian noise and are progressively denoised. This process is analogous to image inpainting tasks, where protein pockets represent the existing parts of an ``image,'' and ligands are the ``missing'' parts that need to be filled in. Current approaches typically handle the ligand either as a whole entity~\cite{guan20233d,schneuing2022structure} or by decomposing ligands into fragments for sampling with pre-imposed priors~\cite{guan2023decompdiff,zhou2023decompopt}. In this work, we apply our guidance strategy to both of these methods.

The idea of diffusion-model-based SBDD is to learn a joint distribution between the protein pocket $P$ and the ligand molecule $M$. The spatial coordinates $x\in \mathbb{R}^{N\times 3}$ and atom features $v\in \mathbb{R}^{N\times K}$ are modeled separately by Gaussian $\mathcal{N}$ and categorical distributions $\mathcal{C}$, respectively, due to their continuous and discontinuous nature. Here $N$ is the number of atoms and $K$ is the number of element types. The forward diffusion process is defined as follows~\cite{guan20233d,schneuing2024structure}:

\begin{equation}
q(M_{t}|M_{t-1},P) = \mathcal{N}(x_{t};\sqrt{1 - \beta_{t}} x_{t-1},\beta_t \textbf{I}) \cdot \mathcal{C}(v_t|(1 - \beta_t)v_{t-1} + \beta_{t}/K).
\end{equation}
Here, $t$ is the timestep and ranges from $0$ to $T$, and $\beta_t$ is the time schedule derived from a sigmoid function. Let $\alpha_t = 1 - \beta_t$ and $\Bar{\alpha}_t = \prod^{t}_{s = 1} \alpha_s$. The reverse diffusion sampling process for spatial coordinates $x$ and atom features $v$ can be done through following Markov chains:
\begin{equation}
\label{VPSDE_iter}
x_{t-1} = \frac{1}{\sqrt{1 - \beta_t}}(x_t + \beta_t \nabla \log_{x_t} P(x_t) ) +\sqrt{\beta_t} z.
\end{equation}

\begin{equation}
\label{bernolli_MC}
v_{t-1} = \arg\max (\widetilde{c}_t(v_t,v_0)),
\end{equation}

where $P(x_t) = \mathcal{N}(x_t;\sqrt{\Bar{\alpha}_t} x_0,(1-\Bar{\alpha}_t)I)$ and $z \sim \mathcal{N}(0,I)$,  
$\widetilde{c}_t(v_t,v_0) = c^*(v_t,v_0)/\sum_{k=1}^{K} c^*_k$, where $c^*(v_t,v_0) = [\alpha_t v_t +(1-\alpha_t)/K]\odot[\Bar{\alpha}_{t-1}v_0+(1-\Bar{\alpha}_{t-1})/K]$.

\subsection{Guidance}\label{guidance}
Guidance is a key advantage of diffusion models, enabling them to model a conditional distribution $P(x_t|y)$ instead of $P(x_t)$, where $y$ is the desired condition for sampled data. In the context of SBDD, $y$ can be the binding affinity $\Delta G$, or other properties like QED and SA.
\subsubsection{Classifier Guidance}
Classifier guidance~\cite{dhariwal2021diffusion} is a plug-and-play method that is straightforward to implement to fine-tune diffusion sampling. It involves decomposing a conditional distribution $P(x_t|y)$ into an unconditional distribution $P(x_t)$ and a classifier term $P(y|x_t)$ through Bayes' Rule:
\begin{equation}\label{bayes_decomp}
    P(x_t|y) = \frac{P(x_t) P(y|x_t)}{P(y)} \propto P(x_t) P(y|x_t).
\end{equation}
To understand classifier guidance, consider that we are interested in maximizing the likelihood that the sampled $x_0$ belongs to class $y$. From a score-matching perspective~\cite{song2019generative, song2020score}, the gradient of the log probability $P(x_t|y)$ with respect to $x_t$ is approximated and simplified through the following steps:
\begin{eqnarray}
    \nabla_{x_t}\log P(x_t|y) = \nabla_{x_t}\log P(x_t)P(y|x_t), \\
    = \nabla_{x_t}\log P(x_t) + \nabla_{x_t}\log P(y|x_t).
\end{eqnarray}
The unconditional gradient term $\nabla_{x_t}\log P(x_t)$
is parameterized by a denoising score network, and $\nabla_{x_t}\log P(y|x_t)$ is modeled by a separately trained classifier. A scaling factor $s$ is then added to control the strength of guidance, and we reach the final expression for classifier guidance:
\begin{equation}\label{3}
    \nabla_{x_t}\log P(x_t|y) = \nabla_{x_t}\log P(x_t) + s\nabla_{x_t}\log P(y|x_t).
\end{equation}

\subsubsection{Classifier-Free Guidance}
Classifier-free guidance further decomposes the term $\nabla_{x_t}\log P(y|x_t)$ back to unconditional score and conditional score, with the final equation being a linear combination of two scores. The scale factor, $s$, is the same scale factor as the one in classifier guidance. The derivation is shown in \eref{clsfree_eqn}: 
\begin{eqnarray}\label{clsfree_eqn}
    \nabla_{x_t}\log P(x_t|y) = \nabla_{x_t}\log P(x_t) + s\nabla_{x_t}\log P(y|x_t), \\
    = \nabla_{x_t}\log P(x_t) + s(\nabla_{x_t}\log P(x_t|y)-\nabla_{x_t}\log P(x_t)), \\
    = (1-s)\nabla_{x_t}\log P(x_t) + s\nabla_{x_t}\log P(x_t|y).
\end{eqnarray}
Classifier-free guidance eliminates the need for separately training a classifier by decomposing the $P(y|x_t)$ term back into two noise terms that can be parametrized by the diffusion model.

\section{Methods}\label{method_major}
\begin{figure}[ht]
  \centering
  \includegraphics[width=\textwidth]{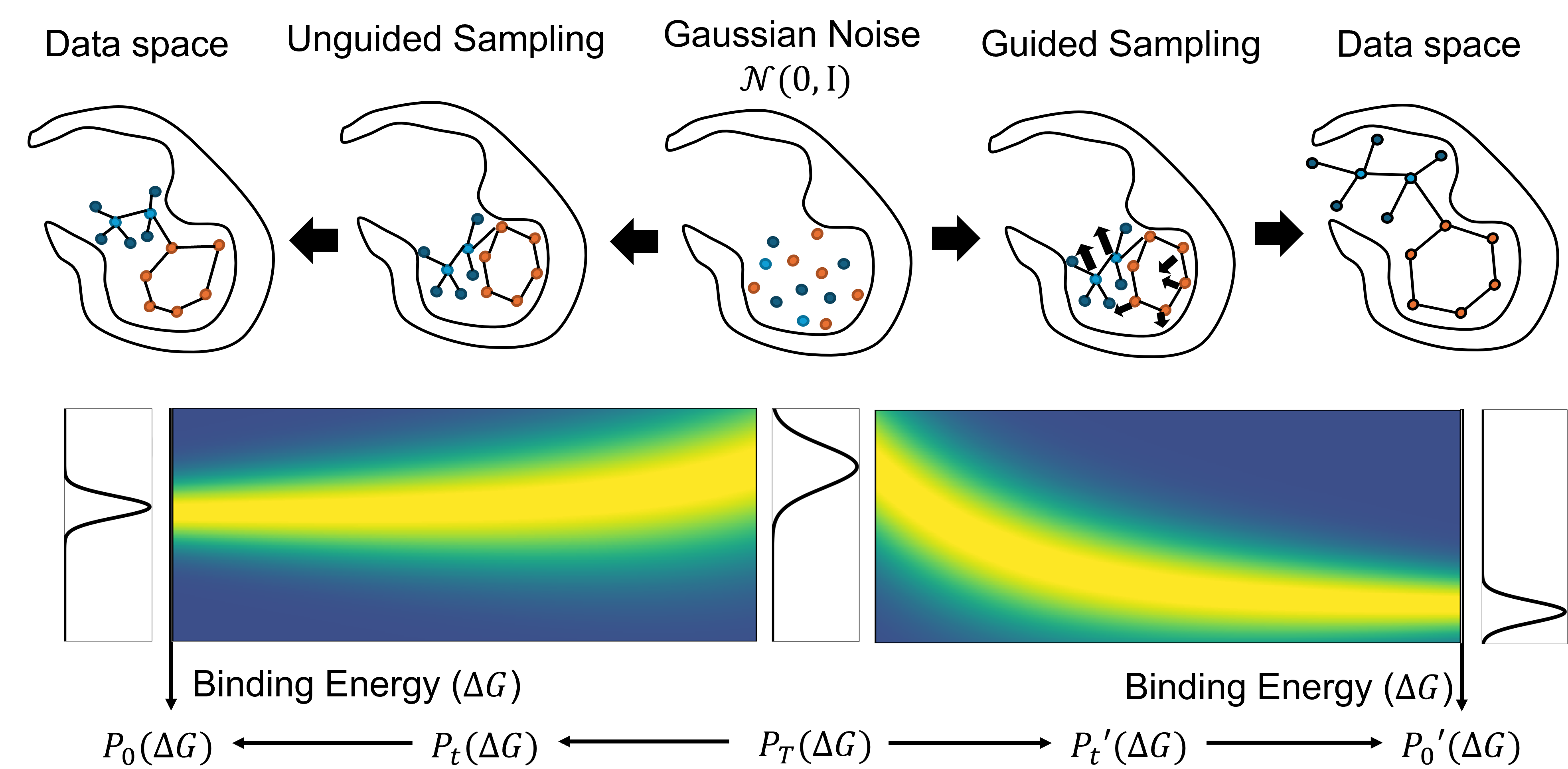}
  \caption{\textbf{Overview of \modelns, a general guidance framework for diffusion-based molecular generation.}
    Top: unguided (left) vs. guided (right) sampling trajectories from Gaussian noise to molecular structures. \model (right) employs either classifier guidance (gradient-based refinement using a trained classifier) or classifier-free guidance (mixture of conditional and unconditional noise predictions). Bottom: evolution of binding energy distributions $P_t(\Delta G)$, showing that guided sampling (right) under \model shifts samples toward lower binding energies.}
  \label{mainfig}
\end{figure}

We introduce the main components of our method: \modelnobf is a versatile, easy-to-use diffusion guidance method for improving ligand-protein pocket binding affinity and other properties in SBDD. We include a schematic in~\fref{mainfig}. \modelnobf consists of two variants:

(1) \textbf{Classifier Guidance.} Classifier guidance guides the diffusion model with a separately trained classifier designed for predicting continuous scalar properties. The classifier is architecture agnostic and can be used with any pre-trained diffusion model. We also explore guiding with multiple constraints within this framework (see~\sref{method:classifier guidance}).

(2) \textbf{Classifier-Free Guidance.} In contrast to Classifier Guidance, Classifier-Free Guidance does not require training a separate classifier network. Instead, it incorporates the property condition directly into the diffusion model training process. Specifically, during training, the condition is randomly concatenated with the input to the score network. During sampling, guidance is achieved by linearly combining the conditional and unconditional score estimates produced by the same diffusion score network. In the context of SBDD, the condition is the binding affinity. 

We also provide detailed descriptions of the datasets and baseline models in~\sref{subsec:dataset-model}.

\subsection{Classifier Guidance}
\label{method:classifier guidance}
\subsubsection{Parameterized Gaussian Distribution for Continuous Property Condition Modeling}
The key aspect of classifier guidance is to model the conditional distribution $P(x_t|y)$ through modeling an extra term $P(y|x_t)$, which reflects the probability distribution of condition $y$ given the input data $x_t$. In image generation, $y$ usually represents the class belonging to a sampled image $x_t$, e.g., flower, dog, etc., which are discretely distributed. In this case, $P(y|x_t)$ represents the probability mass of a certain class and the neural network is learning a $y$ that is Bernoulli distributed. However, in SBDD, $y$ is usually the properties of the ligand, e.g., binding affinity, QED, SA, etc. Unlike the image class label that follows a multinomial distribution, these properties are continuously distributed within certain intervals, and the probability density of $P(y|x_t)$ cannot explicitly be predicted by the NN without an appropriate prior distribution. To resolve this issue, we propose to use a Gaussian as a prior distribution to model $P(y|x_t)$, because these properties are usually normally distributed.

Formally, for a property $y$, if the our desired value is $y = c$, we will then use following Gaussian distribution to model $P(y|x_t)$:
\begin{equation}\label{energy_func_gaussian}
    P(y|x_t) = \frac{1}{\sigma \sqrt{2\pi}}\exp(-\frac{1}{2}\frac{(y_\theta(x_t) - c)^2}{\sigma^2}),
\end{equation}
where $x_t$ is the sampled ligand molecule and $y_\theta(x_t)$ is a property classifier for predicting $y$ given input $x_t$. The intuition behind this is to use the gradient of a Gaussian distribution with its mean set to $c$. By doing so, this can steer the property $y$ of the sample $x_t$ towards the region that is close to $c$. Specifically, $c$ is the value of the binding affinity. In SBDD, a low binding affinity typically around $-5 \sim -10  \text{kcal/mol}$ is desired, and we can set $c$ to be some value in this range and guide the sampled molecule to have a low binding affinity with the protein pocket.

Finally, by substituting \eref{energy_func_gaussian} back to \eref{3}, we have following expression for the conditional score $\nabla_{x_t} \log P(x_t|y)$,
\begin{equation}
\label{guidance_grad_expression}
    \nabla_{x_t}\log P(x_t|y) = \nabla_{x_t}\log P(x_t) - (\frac{s}{2\sigma^2})\nabla_{x_t}(y_\theta(x_t) - c)^2.
\end{equation}
The gradient term is simply reduced into a Mean Squared Error term between the network predicted property value of $x_t$ and the desired value $c$. When guiding on binding affinity, $c$ is the desired value of the binding affinity that we want to guide the ligand to, and can be denoted as $\Delta G_{target}$. We provide the full derivation in~\supsref{1}. 

In our case, $y_\theta(x_t)$ is the binding affinity output from the classifier given the input protein-ligand pair and is denoted as $\Delta G_{predict}([x_t,v_t],P)$. We also find that training the binding affinity classifier with ``clean'' data, denoted as $\Delta G_{predict}([x_0,v_0],P)$, can actually boost the guidance performance and we provide a detailed discussion in the~\supsref{17}. We provide a full algorithm for training classifier in~\supsref{3} and sampling with the classifier in~\supsref{4}. Moreover, we also discuss the details of architecture choices for the classifiers used in~\supsref{7}.

\subsubsection{Generalization of the Energy Function}
In~\eref{energy_func_gaussian}, we use a Gaussian distribution with mean at $c$ to model the conditional term $P(y|x_t)$. We show that this characterization distribution can be generalized and $P(y|x_t)$ can take other bell-shaped distributions with a mean at $c$. Specifically, we can use a synthesized exponential distribution to model $P(y|x_t)$ which is shown in~\eref{exponential},

\begin{equation}
\label{exponential}
P(y|x_t) = \lambda \exp(-\lambda|y_\theta(x_t)-c|).
\end{equation}

Then, the conditional score term will be expressed in the form shown in~\eref{exponential_decomposition},
\begin{equation}
\label{exponential_decomposition}
    \nabla_{x_t}\log P(x_t|y) = \nabla_{x_t}\log P(x_t) - \lambda \nabla_{x_t}|y_\theta(x_t) - c|.
\end{equation}
Since both Gaussian and exponential distributions yield a similar loss function like term between $y_{\theta}(x_t)$ and $c$, which are $|y_{\theta}(x_t) - c|$ and $(y_{\theta}(x_t) - c)^2$, we use a general notation $\mathcal{L}(y_\theta,c)$ to denote this loss function. We rewrite~\eref{guidance_grad_expression} as ~\eref{guidance_grad_expression_general} as below:
\begin{equation}
\label{guidance_grad_expression_general}
\nabla_{x_t}\log P(x_t|y) = \nabla_{x_t}\log P(x_t) - \omega\nabla_{x_t}\mathcal{L}(y_\theta,c),
\end{equation}
where $\omega > 0$. Full ablations of this method are in~\supsref{2}.

\subsubsection{Guidance Strategy}

In order to draw samples from a distribution conditioned on $y$, we need to modify the unconditional Langevin dynamics in~\eref{VPSDE_iter} into the format shown in~\eref{VPSDE_condition},
\begin{equation}
\label{VPSDE_condition}
x_{t-1} = \frac{1}{\sqrt{1 - \beta_t}}(x_t + \beta_t \nabla_{x_t} \log P(x_t|y) ) +\sqrt{\beta_t} z.
\end{equation}
By plugging in ~\eref{guidance_grad_expression}, we reach the iteration rule shown in~\eref{VPSDE_guide}:
\begin{equation}
\label{VPSDE_guide}
x_{t-1} = \frac{1}{\sqrt{1 - \beta_t}}(x_t + \beta_t \nabla_{x_t} \log P(x_t) ) - \textcolor{blue}{\frac{\beta_t}{\sqrt{\alpha_t}} \omega \nabla_{\boldsymbol{x_{t}}}\mathcal{L}(\Delta G_{predict},\Delta G_{target})} +\sqrt{\beta_t} z.
\end{equation}
\eref{VPSDE_guide} is the key equation for the classifier-guided sampling rule. The guidance term helps with steering the distribution toward the desired region during sampling, highlighted in blue. Note that $\Delta G_{predict}$ can either be $\Delta G_{predict}([x_t,v_t],P)$ or $\Delta G_{predict}([x_0,v_0],P)$. There are two complementary ways to understand the guidance term: (1) from an optimization perspective, the additional term can be interpreted as steering the sample $x_t$ to minimize the loss function, effectively guiding it toward regions of higher target likelihood; and (2) from a diffusion model perspective, the added term is part of the conditional score $\nabla_{x_t}\log P(x_t \mid y)$, which enables the sampling process to reconstruct data consistent with the conditional distribution $P(x_t \mid y)$.

In diffusion models, the score term $\nabla_{x_t}\log P(x_t)$ can either be parameterized in noise $\epsilon_t$ space or in $x_0$ space. We follow~\citet{guan20233d} and parameterize the score term in $x_0$ space, as this yields better performance when training an unconditional diffusion model for SBDD. The parametrization is shown in~\eref{parametrization_ofpx}:

\begin{equation}
\label{parametrization_ofpx}
\nabla_{x_t}\log P(x_t) = -\frac{1}{1 - \Bar{\alpha}_t} x_t + \frac{\sqrt{\Bar{\alpha}_t}}{1 - \Bar{\alpha}_t}\hat{x}_0([x_t,v_t],P).
\end{equation}

\subsubsection{Multi-Constraint Classifier Guidance}
\label{method:multiconstraints classifier guidance}
In addition to classifier guidance on binding affinity, our framework can also be utilized to achieve multi-objective guidance, which allows for the optimization of properties beyond just binding affinity scores. We train a multi-constraint prediction model in a similar fashion as classifier guidance on binding affinity. We use a multi-constraint output head and a weighted loss term combining Binding Affinity, Quantitative Drug Likeness (QED), and Synthetic Accessibility (SA) for training, as formulated in~\eref{eqn:multi-train-loss}. Here, $\hat{y}$ denotes a predicted value of $y$ by the regression network:
\begin{equation}
\label{eqn:multi-train-loss}
    \mathcal{L}_{total} = w_{vina}\cdot \mathcal{L}(\hat{\Delta G}_{predict}, \Delta G_{target}) + w_{QED}\cdot \mathcal{L}(\hat{QED}, QED) + w_{SA}\cdot \mathcal{L}(\hat{SA}, SA).
\end{equation}
We can then perform multi-constraint guided sampling, as in~\eref{eqn:multi-sample-loss}:
\begin{equation}
\label{eqn:multi-sample-loss}
x_{t-1} = \frac{1}{\sqrt{1 - \beta_t}}(x_t + \beta_t \nabla_{x_t} \log P(x_t) ) - \frac{\beta_t}{\sqrt{\alpha_t}} \omega \nabla_{\boldsymbol{x_{t}}}\mathcal{L}_{total} +\sqrt{\beta_t} z.
\end{equation}
We show multi-constraint guidance results alongside Binding Affinity Diffusion Guidance in ~\tref{sample-table}. The details on specific weightings used to train the multi-constraint regression model and sampling with the model are provided in~\supsref{7}.

\subsection{Classifier-Free Guidance}
\label{method:classifier free guidance}
\subsubsection{Guided Sampling}

For classifier-free guidance, we substitute~\eref{clsfree_eqn} into~\eref{VPSDE_condition} to get~\eref{clsfree_langevin}. The intuition for this equation is that at each time step, the conditional score term $\nabla_{x_t}\log P(x_t|y)$ will be modeled by a linear combination of both the conditional and unconditional score:
\begin{equation}
\label{clsfree_langevin}
x_{t-1} = \frac{1}{\sqrt{1 - \beta_t}}(x_t + \beta_t ((1-s)\nabla_{x_t}\log P(x_t) + s\nabla_{x_t}\log P(x_t|y))) +\sqrt{\beta_t} z.
\end{equation}

We provide a full algorithm for classifier free sampling in~\supsref{5}.

\subsubsection{Training of Guided Diffusion Model}
We discuss the training and parameterization of the diffusion model in the classifier-free guidance setting. We still parameterize the diffusion in $x_0$ space similar to~\eref{parametrization_ofpx}, for both the conditional and unconditional model. Both $\nabla_{x_t}\log P(x_t)$ and $\nabla_{x_t}\log P(x_t|y)$ are predicted and calculated with the same network shown in~\eref{joint_predict}:
\begin{equation}
\label{joint_predict}
\begin{cases}
\nabla_{x_t}\log P(x_t) =  -\frac{1}{1 - \Bar{\alpha}_t} x_t + \frac{\sqrt{\Bar{\alpha}_t}}{1 - \Bar{\alpha}_t}\hat{x}_0([x_t,v_t],y = \varnothing,P)\\
\nabla_{x_t}\log P(x_t|y=c) = -\frac{1}{1 - \Bar{\alpha}_t} x_t + \frac{\sqrt{\Bar{\alpha}_t}}{1 - \Bar{\alpha}_t}\hat{x}_0([x_t,v_t],y = c,P)
\end{cases}
\end{equation}

We train the diffusion model by randomly discarding the condition $y$ input to be a $\varnothing$ with a certain set probability. The $\varnothing$ here is set to be a manually picked scalar value as a hyperparameter. We assign $\varnothing$ to keep dimension consistency for the input data so that we can jointly train the conditional and unconditional score network together. The training algorithm is shown in~\supsref{6}.

\subsection{Dataset and Model Baselines}\label{subsec:dataset-model}

\paragraph{Dataset.}

We test the methods on two benchmark datasets: \textbf{CrossDocked2020}~\cite{francoeur2020three} and \textbf{PDBBind v2020}~\cite{wang2005pdbbind}. For CrossDocked2020, our data preprocessing and splitting procedures follow the same setting used in TargetDiff and DecompDiff~\cite{guan20233d,guan2023decompdiff}. Following~\citet{guan20233d}, we filter 22.5 million docked protein-ligand complexes based on the criteria of low RMSD for the selected poses (< 1 \AA) and sequence identity less than $30\%$. We select 100,000 complexes for training and 100 complexes for testing. For training the classifier used for guidance, both the previous training complexes and the test complexes are included for training. For evaluation, we sample 100 ligands from each pocket, resulting in a total of 10,000 ligands sampled for benchmarking.

To evaluate our method on higher-quality data, we also conduct experiments on \textbf{PDBBind v2020} and \textbf{CASF2016}\cite{CASF}. Since CASF2016 is a subset of PDBBind, we use \textbf{CASF2016} as the test set and the remaining portion of PDBBind v2020 for training. We clean the dataset by retaining only the entries whose elemental compositions are supported by AutoDock Vina, resulting in 12,126 training samples and 184 test samples.

\paragraph{Baselines.}
We benchmark the performance of our guidance methods on two state-of-the-art diffusion models for SBDD: \textbf{TargetDiff}~\cite{guan20233d} and \textbf{DecompDiff}~\cite{guan2023decompdiff}. We benchmark the classifier guidance variant on \textbf{TargetDiff} and \textbf{DecompDiff} to show its plug-and-play feature. We then benchmark the classifier-free guidance variant on \textbf{TargetDiff}. For \textbf{DecompDiff}, we experiment with two types of priors used in their paper: the reference prior, which we denote as \textbf{DecompDiff Ref}, and the pocket prior, which we denote as \textbf{DecompDiff Beta}. 
We additionally include \textbf{DiffSBDD}~\cite{schneuing2024structure}, a recently proposed diffusion-based SBDD method that jointly models atomic coordinates and molecular topology.
We include two other SBDD diffusion models as baselines: \textbf{IPDiff}~\cite{huang2024proteinligand}, and \textbf{BindDM}~\cite{huang2024binding}. We also compare \model with \textbf{DecompOpt}~\cite{zhou2023decompopt}, an optimization method built for diffusion models for SBDD. Specifically, for \textbf{DecompOpt}, we select the groups in~\citet{zhou2023decompopt}: TargetDiff + Optimization, which we denote as \textbf{TargetDiff w/ Opt.}, and DecompDiff + Optimization, which we denote as \textbf{DecompOpt}. We also compare our results with non-diffusion SBDD models: \textbf{liGAN}~\cite{ragoza2022generating}, \textbf{GraphBP}~\cite{liu2022generating}, \textbf{AR}~\cite{luo20213d}, \textbf{Pocket2Mol}~\cite{peng2022pocket2mol}.

\section{Results and Discussion}
\label{sec:results}

We present and analyze the findings revealed by \model, beginning with improvements in binding affinity and selectivity in~\sref{subsec:classifier guide} and~\sref{specificity}, then examining geometric pose quality in~\sref{posequal}, and concluding with practical implications and limitations in~\sref{tradeoff} and~\sref{limitation}. Throughout this section, we refer to \textbf{Classifier Guidance (CG)}, \textbf{Classifier-Free Guidance (CFG)}, and \textbf{Multi-Constraints Classifier Guidance (MC-CG)} as our three guidance variants.

\subsection{Improvements in Binding Affinity and Molecular Properties}
\label{subsec:classifier guide}

\begin{table}[h]
  \caption{Summary table of binding affinity performance and other properties for different guidance methods on top of a base generative model. For each metric, the top two methods are highlighted---\textbf{bolded} for the first and \underline{underlined} for the second. The methods are categorized into three groups: non-Diffusion methods (\textbf{non-Diff.}), Diffusion methods (\textbf{Diff.}), and Diffusion methods with BADGER (\textbf{Diff. + BADGER}).}
  \label{sample-table}
  \centering
  \begin{adjustbox}{width=\columnwidth,center}
  \begin{tabular}{c|c|c c|c c|c c|c c|c c|c c|c}
    \toprule
    
    \multicolumn{2}{c}{Metric}  & \multicolumn{2}{c}{Vina Score$\downarrow$} & \multicolumn{2}{c}{Vina Min$\downarrow$}  & \multicolumn{2}{c}{QED$\uparrow$} & \multicolumn{2}{c}{SA$\uparrow$} & \multicolumn{2}{c}{Diversity$\uparrow$} & \multicolumn{2}{c}{High Affinity(\%)$\uparrow$} & Specificity Score$\downarrow$ \\
    \multicolumn{2}{c}{Group name}   & Mean ($\Delta\%$)   & Med. ($\Delta\%$)  & Mean ($\Delta\%$) & Med. ($\Delta\%$)&Mean & Med. & Mean & Med. & Mean & Med. &Mean & Med. & Mean\\
    
    \midrule
    
    \multicolumn{2}{c}{Ref.} & -6.36 & -6.46 & -6.71 & -6.49& 0.48& 0.47 & 0.73& 0.74& -&-&-&- & - \\
    
    \midrule
    
    \multirow{6}{*}{non-Diff.}& liGAN\cite{ragoza2022generating} & - & - & - & -& 0.39& 0.39 & 0.59& 0.57 &0.66&0.67&21.1&11.1 & -1.46 \\
    &GraphBP\cite{liu2022generating} & - & - & - & -& 0.43& 0.45 & 0.49& 0.48 &0.79&0.78&14.2&6.7 & - \\
    &AR\cite{luo20213d} & -5.75 & -5.64 & -6.18 & -5.88& 0.51& 0.50 & 0.63& 0.63 & 0.70 & 0.70 & 37.9&31.0 & -1.68 \\
    &Pocket2Mol\cite{peng2022pocket2mol} & -5.14 & -4.70 & -6.42 & -5.82& \textbf{0.56} & \textbf{0.57} & \textbf{0.74}& \textbf{0.75} &0.69&0.71& 48.4&51.0 & -1.56 \\
    \midrule
    
    \multirow{6}{*}{Diff.}
    & IPDiff\cite{huang2024proteinligand} & -6.42 & -7.01 & -7.45 & -7.48 & 0.52 & 0.53 & 0.61 & 0.59 &0.74& 0.73 & 69.5 & 75.5 & - \\
    & BindDM\cite{huang2024binding} & -5.92 & -6.81 & -7.29 & -7.34  & 0.51 & 0.52 & 0.58 & 0.58 &0.75&0.74&64.8&71.6 & - \\
    & \cellcolor{gray!25}TargetDiff\cite{guan20233d} &  -5.47 & -6.30 & -6.64 & -6.83 & 0.48& 0.48 & 0.58& 0.58 &0.72&0.71&58.1&59.1 & -2.77 \\
    &\cellcolor{gray!25}DecompDiff Ref\cite{guan2023decompdiff} & -4.97 & -4.88 & -6.07 & -5.79 & 0.45 & 0.45 & \underline{0.64}& \underline{0.63}& \textbf{0.82} &\textbf{0.84}&64.6&75.5 & -1.48 \\
    &\cellcolor{gray!25}DecompDiff Beta\cite{guan2023decompdiff} & -4.20 & -5.90 & -6.78 & -7.32 & 0.28 & 0.25 & 0.52 & 0.52 & 0.67 & 0.67 & \underline{77.3} & \underline{94.5} & -2.82 \\
    & \cellcolor{gray!25}DiffSBDD\cite{schneuing2024structure} & -1.01 & -4.54 & -4.34 & -5.51 & 0.46 & 0.47 & 0.57 & 0.57 & 0.74 & 0.75 & 49.8 & 49.0 & -1.89 \\
    
    \midrule
    
    \multirow{2}{*}{Diff. +}
    &TargetDiff + Classifier Guidance & \textbf{-7.70} (+40.8\%) & \textbf{-8.53} (+35.4\%) & \underline{-8.33} (+25.5\%) & \underline{-8.44} (+23.6\%)& 0.46& 0.46& 0.50 & 0.49 &0.78&0.80& 70.2&76.8 & \textbf{-4.28} \\
    &DecompDiff Ref + Classifier Guidance & -6.05 (+21.7\%) & -6.02 (+23.3\%) & -6.76 (+11.4\%) & -6.53 (+12.8\%) & 0.45 & 0.45 & 0.61& 0.60 & \underline{0.81} & \underline{0.82} & 70.8 & 75.8 & -1.51 \\
    &DecompDiff Beta + Classifier Guidance & \underline{-6.72} (+60.0\%) & \underline{-7.96} (+34.9\%) & \textbf{-8.44} (+24.5\%) & \textbf{-8.77} (+19.8\%) & 0.29 & 0.26 & 0.49 & 0.49 & 0.66 & 0.66 &\textbf{83.0}&\textbf{97.8} & \underline{-3.16} \\
    \multirow{2}{*}{BADGER}
    &TargetDiff + Multi-Constraints Classifier Guidance & -6.58 (+20.3\%) & -7.33 (+16.3\%) & -7.55 (+13.7\%) & -7.66 (+12.2\%)& \underline{0.52}& \underline{0.53}& 0.59 & 0.58 &0.79&0.81& 65.3&70.0 & -2.82 \\
    &TargetDiff + Classifier-Free Guidance & -6.18 (+12.9\%) & -6.84 (+8.6\%) & -6.87 (+3.5\%) & -6.92 (+1.3\%)& 0.50& 0.50& 0.58 & 0.58 & 0.79 & 0.80 & 57.5& 56.8 & -2.68 \\
    
    \bottomrule
  \end{tabular}
  \end{adjustbox}
\end{table}

\begin{table}[h]
  \caption{Performance summary for \model on the PDBBind dataset (CASF is a subset of PDBBind and is held out as test dataset, the model and classifier are trained on the remaining PDBBind dataset (\textbf{PDBBindv2020-remain}) for TargetDiff and DecompDiff variants. For each metric, the best and second-best are highlighted: \textbf{bold} for best and \underline{underline} for second-best.}
  \label{pdbbind-table}
  \centering
  \begin{adjustbox}{width=0.99\columnwidth,center}
  \begin{tabular}{c|cc|cc|cc|cc|cc|cc|c|c}
    \toprule
    Metric & \multicolumn{2}{c}{Vina Score$\downarrow$} & \multicolumn{2}{c}{Vina Min$\downarrow$} & \multicolumn{2}{c}{QED$\uparrow$} & \multicolumn{2}{c}{SA$\uparrow$} & \multicolumn{2}{c}{Diversity$\uparrow$} & \multicolumn{2}{c}{High Affinity (\%)$\uparrow$} & \multicolumn{1}{c}{Specificity$\downarrow$} & Structural Validity (\%)$\uparrow$ \\
    Method & Mean & Med. & Mean & Med. & Mean & Med. & Mean & Med. & Mean & Med. & Mean & Med. & Mean & Value \\
    \midrule
    Unguided TargetDiff & -5.90 & -6.40 & -7.20 & -7.15 & 0.46 & 0.47 & \underline{0.57} & \underline{0.56} & \textbf{0.75} & \textbf{0.75} & 37.29 & 33.33 & -0.79 & \textbf{98.70} \\
    Unguided DecompDiff & -6.80 & -6.71 & -7.79 & -7.70 & 0.40 & 0.38 & \underline{0.57} & \underline{0.56} & \textbf{0.75} & \textbf{0.75} & 35.30 & 30.00 & -0.72 & 91.07 \\
    TargetDiff + Classifier Guidance & -6.44 & -6.61 & -7.40 & -7.34 & 0.46 & 0.46 & 0.56 & 0.55 & \underline{0.74} & \underline{0.74} & 40.56 & 37.50 & -1.15 & \underline{98.26} \\
    TargetDiff + Classifier-Free Guidance & \underline{-8.13} & \underline{-8.21} & \underline{-8.50} & \underline{-8.36} & \textbf{0.50} & \textbf{0.52} & 0.54 & 0.53 & \textbf{0.75} & \textbf{0.75} & \underline{52.52} & \underline{52.77} & \underline{-1.80} & 96.47 \\
    DecompDiff + Classifier Guidance & -7.28 & -7.13 & -8.10 & -7.99 & 0.40 & 0.39 & 0.56 & 0.55 & \underline{0.74} & \underline{0.74} & 42.20 & 37.50 & -0.96 & 91.87 \\
    DecompDiff + Classifier-Free Guidance & \textbf{-8.59} & \textbf{-8.55} & \textbf{-9.08} & \textbf{-9.01} & \underline{0.48} & \underline{0.49} & \textbf{0.58} & \textbf{0.57} & \underline{0.74} & \underline{0.74} & \textbf{67.63} & \textbf{76.38} & \textbf{-1.87} & 96.33 \\
    \bottomrule
  \end{tabular}
  \end{adjustbox}
\end{table}

\begin{figure*}[h]
    \centering
    \includegraphics[width=0.98\linewidth]{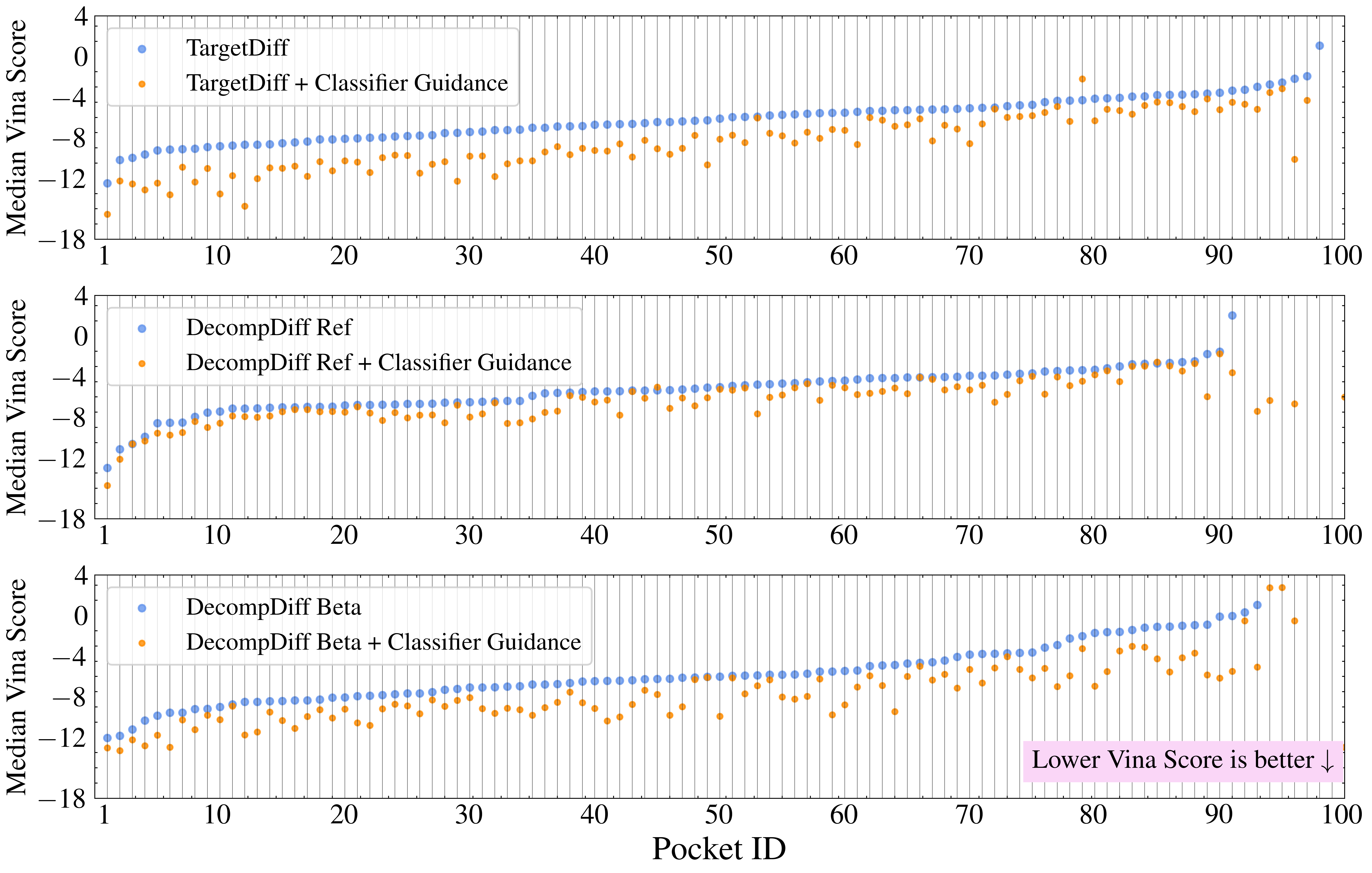}
  \caption{\textbf{Improvement in median Vina Scores across 100 protein pockets after applying Classifier Guidance in \modelns}. Each panel corresponds to a diffusion model variant: TargetDiff (top), DecompDiff Ref (middle), and DecompDiff Beta (bottom). For each pocket, we compare the median Vina Score before (blue) and after (orange) applying classifier-guided sampling. Across all models, \model consistently improves binding quality—achieving lower Vina Scores for $ 99\%$ of the pockets (lower is better $\downarrow$). For a few outlier pockets, the unguided model’s scores exceed the plotted range, yet classifier guidance still yields notable improvements.} 
  \label{pocketwise}
\end{figure*}

\begin{figure*}[!ht]
    \centering
    \includegraphics[width=0.98\linewidth]{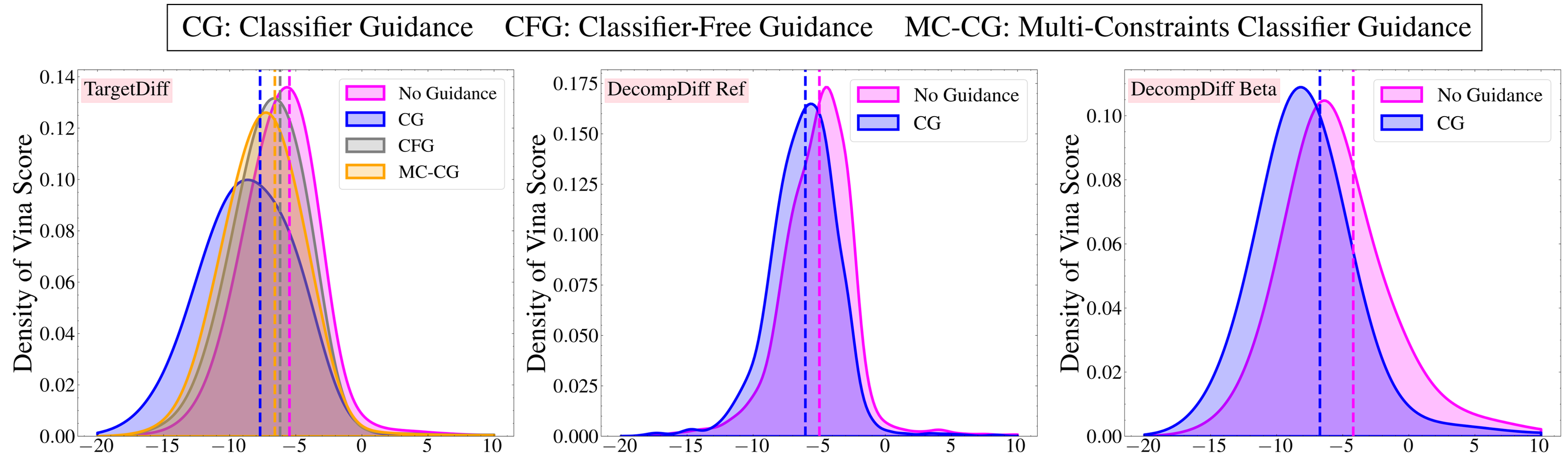}
\caption{\textbf{Distribution of Vina Scores for molecules generated with and without \model guidance across diffusion model baselines.} Shown are results for TargetDiff (left), DecompDiff Ref (middle), and DecompDiff Beta (right). Each plot compares unguided sampling (pink) with guided variants, including Classifier Guidance (CG), Classifier-Free Guidance (CFG), and Multi-Constraint Classifier Guidance (MC-CG). Across all models, \model consistently lowers both the mean and median Vina Scores and shifts the entire distribution toward lower (better) binding energies, potentially suggesting more favorable protein–ligand interactions.} 
  \label{distshift}
\end{figure*}

\textbf{\model improves binding affinity across diffusion backbones.} 
\tref{sample-table} and \tref{pdbbind-table} show that applying BADGER to TargetDiff, DecompDiff Ref, and DecompDiff Beta consistently lowers both Vina Score and Vina Min on both the \textbf{CrossDocked2020} and \textbf{PDBBind v2020} datasets. Both CG and CFG reliably improve affinity across all backbones. Extending CG from a single-constraint formulation to Multi-Constraints Classifier Guidance (MC-CG) further allows BADGER to jointly optimize Vina Score, QED, and SA.

\textbf{Affinity gains are robust across diverse pockets.}  
\fref{pocketwise} visualizes median Vina Scores across 100 pockets from CrossDocked2020. In nearly every case, BADGER shifts the pocket-level median toward stronger (more negative) affinities. Even for challenging pockets where the unguided baseline performs poorly, CG and CFG still yield meaningful improvements. The full affinity distribution in \fref{distshift} further demonstrates that BADGER shifts the entire score distribution—not just summary statistics. This indicates a global improvement in sampled interactions.

\textbf{Molecular properties remain largely preserved.}  
Although BADGER explicitly optimizes for binding affinity, QED and SA remain stable with only modest trade-offs. Because these metrics are typically used as broad filters rather than strict optimization targets, we put less emphasis on QED and SA scores.

\subsection{Binding Specificity Across Protein Pockets}
\label{specificity}

Strong binding alone does not ensure useful molecular design; ligands should also bind selectively to their intended targets. Following \citet{gao2024rethinking}, we identify a top-10 ligand for each pocket and compute its on-target affinity $\Delta G_{\text{on-target}}$. We then cross-dock the same ligand into five randomly selected pockets to obtain off-target affinities $\Delta G_{\text{off-target}}$. The resulting specificity score is:

\begin{equation}
    \text{Specificity Score} = \frac{1}{MN} \sum_{j=1}^{N} \sum_{i_j=1}^{M} \left( \Delta G^{(j)}_{\text{on-target}} - \Delta G^{(i_j)}_{\text{off-target}} \right)
\end{equation}

We present the results in \tref{sample-table} and \tref{pdbbind-table}. Across both CrossDocked2020 and PDBBind v2020 datasets, BADGER improves specificity relative to unguided baselines, with CG and CFG each achieving the best scores in different settings. These results show that BADGER not only strengthens absolute binding affinity but also shifts the generative process toward ligands that preferentially bind to the correct target rather than unrelated pockets.

\subsection{Ligand–Protein Pose Quality}
\label{posequal}

\begin{figure*}[!ht]

      \centering
      \includegraphics[width=\linewidth]{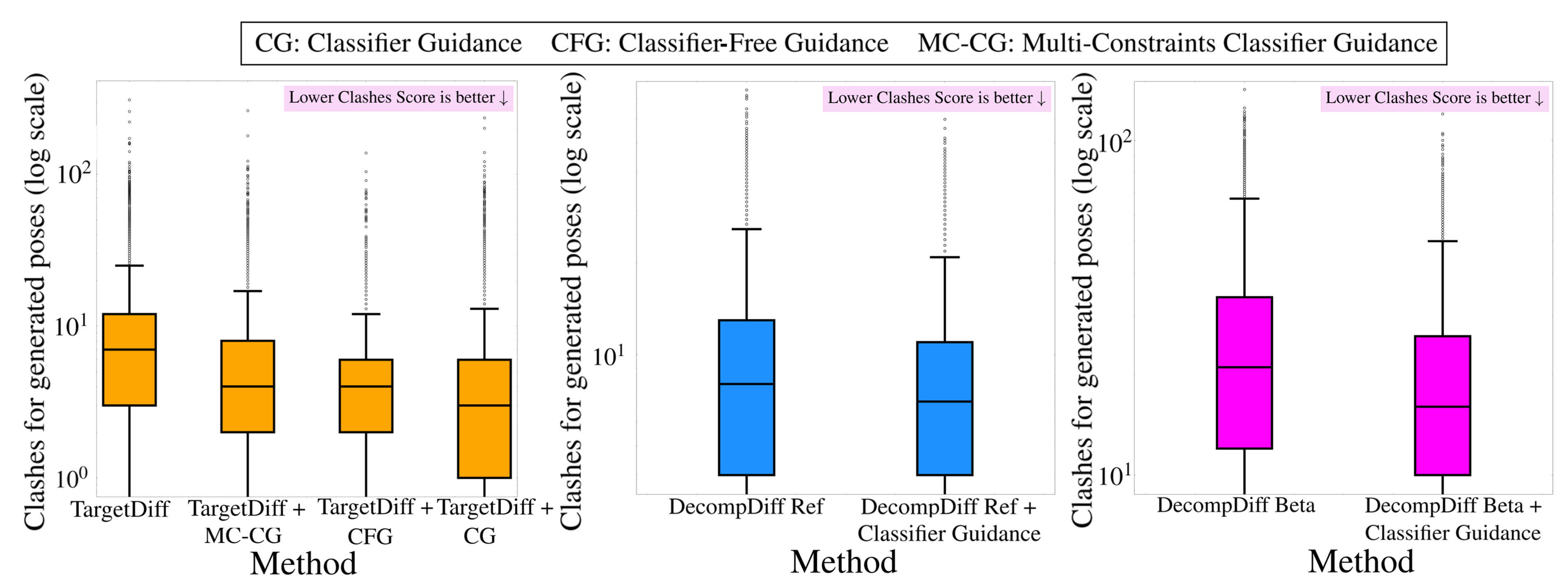}
    \caption{\textbf{Steric Clashes Score improvement with \model across diffusion model variants.} Each box plot reports the distribution (log scale) of steric clashes scores for generated ligand poses reconstructed from sampled molecules. Lower values indicate fewer atomic overlaps and more physically stable conformations. Across all diffusion models—TargetDiff, DecompDiff Ref, and DecompDiff Beta—\model systematically reduces steric clashes under Classifier Guidance (CG), Classifier-Free Guidance (CFG), and Multi-Constraint Classifier Guidance (MC-CG), potentially suggesting improved geometric plausibility of generated poses.}
    \label{clashes}
\end{figure*}

Binding affinity metrics do not capture geometric feasibility. Thus, we additionally evaluate steric clashes following \citet{harris2023benchmarking}. Steric clashes penalize overlapping van der Waals volumes and serve as an indicator of physically unrealistic poses.

\fref{clashes} shows that BADGER reduces steric clashes across all underlying diffusion backbones. The improvement is consistent across CG, CFG, and MC-CG, suggesting that BADGER produces not only higher-affinity ligands but also finds more geometrically plausible poses that better fit the protein pocket.

\subsection{Practical Trade-offs Between Guidance Variants}
\label{tradeoff}

Having examined how BADGER improves binding affinity, specificity, and geometric pose quality, we now discuss the practical considerations that affect which guidance variant is most suitable for a given application. Although CG, CFG, and MC-CG all provide consistent improvements across molecular properties, they differ in computational cost, training stability, and flexibility. Understanding these trade-offs can help practitioners select the guidance strategy best aligned with their computational budget and modeling goals.

Each guidance variant offers advantages under different practical constraints:

\begin{itemize}
\item \textbf{Computational cost:}  
During training, CG requires training a separate classifier. At sampling time, the computational cost depends on the relative sizes of the diffusion model and the classifier. Specifically, for each sampling step, CG involves one forward pass of the diffusion model, one forward pass of the classifier, and one backward pass of the classifier, whereas CFG requires two forward passes of the diffusion model. If the diffusion model is much larger than the classifier, CG can be faster; however, when their sizes are comparable, CFG is typically more efficient.

\item \textbf{Training stability:}  
CG requires training an additional network to predict binding affinity from noisy data. Although we found this process to be stable on the CrossDock2020 and PDBBind datasets, such training may introduce additional uncertainty when scaling to larger models and datasets. In contrast, the CFG requires training only the diffusion model, thereby reducing uncertainty by removing the need to train two separate networks.

\item \textbf{Flexibility:}  
A key advantage of CG is its plug-and-play flexibility. This variant is particularly suitable when a pretrained diffusion model is already available and retraining the diffusion model is computationally more expensive than training a separate classifier.

\end{itemize}

\subsection{Limitations and Future Directions}
\label{limitation}

A primary limitation of this work is the use of docking scores as a proxy for binding affinity, which provides only an approximate measure of true binding strength. Future work should focus on integrating more accurate, experimentally derived binding affinity predictors into the generative guidance process. In practice, this can be achieved by training a classifier on datasets containing experimentally measured binding affinities to provide more reliable guidance signals for the model.
\section{Conclusion}

We introduce \modelnobfns, a guidance method to improve the binding affinity of ligands generated by diffusion models in SBDD. \modelnobf demonstrates that binding affinity awareness can be directly enforced into the sampling process of the diffusion model through either classifier guidance or classifier-free guidance. Our method opens up new avenues for optimizing ligand properties in SBDD. It is also a general method that can be applied to a wide range of datasets and has the potential to better optimize the drug discovery process. 
\section{Data and Software Availability}
The dataset and its processing procedure used in this work can be found at \url{https://github.com/guanjq/targetdiff}.
Our source codes will be made publicly available at \url{https://github.com/ASK-Berkeley/BADGER-SBDD}.

\section{Supporting Information}

This Supporting Information provides derivations, algorithmic details, ablation studies, and extended benchmarking figures and tables that complement the main text.

To assist readers in navigating the supplementary materials, \tref{tab:supp-outline} summarizes the structure of the Supporting Information and briefly describes the purpose of each section.

\begin{table}[h]
\centering
\scriptsize
\caption{Overview of Supporting Information and its role in validating BADGER.}
\label{tab:supp-outline}
\begin{adjustbox}{width=\textwidth,center}
\begin{tabular}{p{4.2cm} p{10.8cm}}
\toprule
Supporting Information Section & Purpose and Relation to Main Text \\
\midrule
\midrule
Derivation of classifier-guidance conditional score &
Provides the full mathematical derivation of the continuous-property conditional score used in BADGER’s classifier guidance (~\sref{method:classifier guidance}), clarifying how the Gaussian prior leads to the gradient term in \eref{VPSDE_guide} and justifying the sampling update. \\
\midrule

Ablation on energy (loss) function &
Evaluates Gaussian vs.\ exponential formulations for the conditional likelihood, demonstrating why the Gaussian MSE-style energy yields more stable and effective guidance (supporting~\sref{method:classifier guidance} and \tref{sample-table}). \\
\midrule

Algorithms for classifier training and classifier-guided sampling &
Provides complete pseudocode (Algorithms 1–2) for training affinity/QED/SA regressors and for guided sampling, ensuring reproducibility of BADGER’s plug-and-play variant and complementing~\sref{method:classifier guidance}. \\
\midrule

Algorithms for classifier-free guidance (training and sampling) &
Gives full procedures for classifier-free training and sampling (Algorithms 3–4), showing how conditional and unconditional scores are combined (complements~\sref{method:classifier free guidance}). \\
\midrule

Implementation details (architectures, hyperparameters, datasets) &
Documents complete architectural choices (EGNN, Transformer), hyperparameters to support reproducibility for all experiments. \\
\midrule

Benchmarking vs.\ DecompOpt &
Compares BADGER to optimization-based post-processing (DecompOpt) under matched sampling conditions, showing that BADGER achieves higher affinity without requiring large candidate pools. \\
\midrule

Additional binding-affinity results (pocket-wise plots) &
Extends \fref{pocketwise} by providing full pocket-level affinity improvements for CG, MC-CG, and CFG across all 100 CrossDocked pockets, showing consistency across pockets. \\
\midrule

Redocking RMSD analyses &
Reports full redocking RMSD distributions, supporting \sref{posequal} and demonstrating that BADGER-generated poses show improved agreement with the Vina scoring function. \\
\midrule

Consistency with experimental $\Delta G$ &
Quantitatively compares Vina energies with experimentally measured $\Delta G$ from PDBBind2016, discussing the limitation of main-text justification for using Vina as a guiding signal (~\sref{method:classifier guidance}). \\
\midrule

Ablation on guidance strength $s$ &
Analyzes how different guidance strengths shift affinity while maintaining chemical validity, helping identify practical $s$ ranges (supports~\sref{method:classifier guidance}). \\
\midrule

Benchmarking structural validity, specificity, bond geometry, and chirality &
Evaluates structural validity, specificity (cross-docking), and geometric realism (bond angles/lengths, chiral centers), showing BADGER maintains chemical plausibility while improving affinity and selectivity. \\
\midrule

Ablation: imperfect classifier and context sweep $c$ &
Studies robustness to classifier quality and sensitivity to conditioning targets. \\
\midrule

Benchmarking sampling speed &
Reports runtime overhead of CG, CFG, and MC-CG relative to unguided diffusion, supporting the practical-efficiency discussion in~\sref{tradeoff}. \\

\midrule
Analysis of poor binding in outlier pockets &
Investigates potential factors underlying binding pockets where unguided diffusion models consistently produce poor affinity. \\

\midrule
Discussion on clean vs.\ noisy classifier training &
Discusses potential reasons why training classifiers on clean data improves guidance performance.  \\

\bottomrule

\end{tabular}
\end{adjustbox}
\end{table}

\section{Acknowledgements}
This work was supported by Laboratory Directed Research and Development (LDRD) funding under Contract Number DE-AC02-05CH11231. We thank Eric Qu, Sanjeev Raja, Toby Kreiman, Rasmus Malik Hoeegh Lindrup and Nithin Chalapathi for their insightful opinions on this work. We also thank Bo Qiang, Bowen Gao, and Xiangxin Zhou for their helpful suggestions on reproducing the benchmark models.

\noindent
\textbf{Funding Sources:} Laboratory Directed Research and Development (LDRD) under Contract No. DE-AC02-05CH11231.

\section*{Conflict of Interest}
\textit{The authors declare no competing financial interests.}

\section*{ORCID}
Yue Jian: https://orcid.org/0000-0003-2464-3337 \\
Curtis Wu: https://orcid.org/0009-0001-3656-8031\\
Danny Reidenbach: https://orcid.org/0000-0002-2973-8709 \\
Aditi S. Krishnapriyan: https://orcid.org/0000-0003-3472-6080

\section*{Author Contributions}
Y.J. and C.W. contributed equally to this work. A.S.K., Y.J., and D.R. conceived the idea. Y.J. led the project, and Y.J., C.W., and D.R. developed it. A.S.K. supervised the project development.

\clearpage
\section*{Supporting Information}

\setcounter{section}{0}
\renewcommand{\thesection}{S\arabic{section}}

\section{Derivation for the classifier guidance variant conditional score expression}\label{energy_derive_sup}
We start from the definition of classifier guidance:
\begin{equation}\label{basic_def}
    \nabla_{x_t}\log P(x_t|y) = \nabla_{x_t}\log P(x_t) + s\nabla_{x_t}\log P(y|x_t).
\end{equation}
We want the sample data $x_t$ to be conditioned on a scalar value $y=c$. We use a Gaussian to model $P(y|x_t)$, because we want most of our sample to fall at around $y=c$. This can be understood as expressing $P(y|x_t)$ to be a Gaussian with mean $c$, which is:
\begin{equation}\label{energy_func_gaussian_sup}
    P(y|x_t) = \frac{1}{\sigma \sqrt{2\pi}}\exp(-\frac{1}{2}\frac{(y_\theta(x_t) - c)^2}{\sigma^2}).
\end{equation}
We then plug \eref{energy_func_gaussian_sup} into \eref{basic_def}, which becomes:
\begin{eqnarray}
\nabla_{x_t}\log P(x_t|y) = \nabla_{x_t}\log P(x_t) + s\nabla_{x_t}\log (\frac{1}{\sigma \sqrt{2\pi}}\exp(-\frac{1}{2}\frac{(y_\theta(x_t) - c)^2}{\sigma^2}))\\
= \nabla_{x_t}\log P(x_t) + s(\cancel{\nabla_{x_t}\log \frac{1}{\sigma \sqrt{2\pi}}} + \nabla_{x_t}\log \exp(-\frac{1}{2}\frac{(y_\theta(x_t) - c)^2}{\sigma^2}))\\
\end{eqnarray}
We group all the terms together with $s$ into a constant $S$ and reach:
\begin{eqnarray}
    \nabla_{x_t}\log P(x_t|y) = \nabla_{x_t}\log P(x_t) - (\frac{s}{2\sigma^2})\nabla_{x_t}(y_\theta(x_t) - c)^2 , \\
    = \nabla_{x_t}\log P(x_t) - S\nabla_{x_t}(y_\theta(x_t) - c)^2.
\end{eqnarray}

\section{Ablation on the type of energy function}\label{ablate loss function section}
We provide a full ablation on different types of loss functions used in~\tref{ablate loss function}.

\begin{table}[h]
    \caption{Ablation on the different types of loss functions for classifier guidance}
    \label{ablate loss function}
    \centering
    \begin{tabular}{c|c c|c c|c c|c c}
        \toprule
         Loss function type& \multicolumn{2}{c}{Vina Score} & \multicolumn{2}{c}{Vina Min} & \multicolumn{2}{c}{QED} &\multicolumn{2}{c}{SA} \\
          &Mean & Med & Mean & Med & Mean & Med& Mean & Med \\
         \midrule
         No guidance& -5.47& -6.30& -6.64& -6.83& 0.48& 0.48& 0.58&0.58\\
         Exponential &-6.06& -6.82& -7.16& -7.20& 0.50& 0.50& 0.59& 0.59\\
         Gaussian &-6.98& -7.57& -7.78& -7.85& 0.50& 0.50& 0.59& 0.58\\
        \bottomrule
    \end{tabular}
    \label{tab:multiprop-table}
\end{table}

\section{Algorithm for training the classifier}\label{alg:clsf_sup}

We outline the full algorithm for training our classifier used in classifier guidance, which is discussed in the\mref 3.1.1.

\begin{algorithm}[H]
\caption{Algorithm for training classifier}\label{alg:train}
\hspace*{\algorithmicindent} \textbf{Input} The protein-ligand binding dataset $\{(P_i,M_i),\Delta G_{i}\}^{N}_{i = 1}$, a neural network $y_{\theta}()$
\begin{algorithmic}
\While{$y_{\theta}()$ does not converge}
\For{$i=$ shuffle $\{1,2,3,4,...,N\}$}
\State Predict binding affinity with network $\Delta \hat{G}_i = y_{\theta}(P_i,M_i)$
\State Calculate MSE loss for binding affinity $\mathcal{L} = ||\Delta \hat{G}_i - \Delta G_i||_2$
\State Mask out loss if the ground truth binding affinity is invalid: $\mathcal{L} \gets 0$ if $\Delta G_i > 0$
\State update $\theta$ base on loss $\mathcal{L}$
\EndFor
\EndWhile

\end{algorithmic}
\end{algorithm}

\section{Algorithm for classifier guidance sampling }\label{alg:clsf_sup_sample}

We outline the full algorithm for our classifier guidance sampling method, which is described in\mref 3.1.

\begin{algorithm}[H]
\caption{Sampling Algorithm for Classifier Guidance}\label{alg_cap}
\hspace*{\algorithmicindent} \textbf{Input} The protein binding pocket $P$, learned diffusion model $\phi_{\theta}$, classifier for binding affinity prediction $f_{\psi}$, target binding affinity $\Delta G_{target}$, scale factor on guidance $s$ \\
\hspace*{\algorithmicindent} \textbf{Output} Sampled ligand molecule $M$ that binds to pocket $P$
\begin{algorithmic}
\State Sample number of atoms in $M$ based on the prior distribution conditioned on pocket size
\State Move the center of mass of protein pocket $P$ to zero, do the same movement for ligand $M$
\State Sample initial molecular atom coordinates $x_T$ and atom types $v_T$
    \State $x_T \in \mathcal{N}(0,\textbf{I})$
    \State $v_T = one\_hot(\arg\max_{i}(g_{i})), where\ g \sim Gumble(0,1)$
\For{t in $T,T-1,...,1$}
    \State {Predict [$\hat{x_0},\hat{v_0}$] through [$\hat{x_0},\hat{v_0}$] = $\phi_{\theta}$([$x_t,v_t$], t, $P$)}
    \State Calculate guidance $g = \nabla_{x_t}||f_{\psi}(P,[\hat{x}_{0},\hat{v}_{0}]) - \Delta G_{target}||_2$ 
    \State $\tilde{\mu}_t(x_t,\hat{x}_0) = \frac{\sqrt{\Bar{\alpha}_{t-1}}\beta_t}{(1-\Bar{\alpha}_t)}\hat{x}_0 + \frac{\sqrt{\alpha_t}(1-\Bar{\alpha}_{t-1})}{(1-\Bar{\alpha}_t)}x_t$
    \State Apply guidance:
        \State $\tilde{\mu}_t'(x_t,\hat{x}_0) = \tilde{\mu}_t(x_t,\hat{x}_0) - s\frac{\beta_t }{\sqrt{\alpha_t}}g$
    \State $\Tilde{\beta}_t = \frac{1-\Bar{\alpha}_{t-1}}{1-\Bar{\alpha}_t}\beta_t$
    \State sample $\epsilon \sim \mathcal{N}(0,\textbf{I})$
    \State $x_{t-1} = \epsilon\sqrt{\Tilde{\beta}_t}+\tilde{\mu}_t'(x_t,\hat{x}_0)$

    \State Sample $v_{t-1}$ from $q_{\theta}(v_{t-1}|v_t, \hat{v}_0) = \mathcal{C}(v_{t-1}|\Tilde{c}(v_t,v_0))$
    \State $\Tilde{c}(v_t,v_0)=(\alpha_t v_t+(1-\alpha_t)/K)\odot(\Bar{\alpha}_{t-1}v_0+(1-\Bar{\alpha}_{t-1})/K)$
    \State Sample $v_{t-1}$
    \Indent
    \State $v_{t-1} = \arg\max(\Tilde{c}(v_t,v_0))$
    \EndIndent
\EndFor

\end{algorithmic}
\end{algorithm}

\section{Algorithm for classifier-free guidance sampling}\label{alg:clsf_free_sample_sup}

We outline the full algorithm for our classifier-free guidance sampling method, which is described in\mref 3.2.

\begin{algorithm}[H]
\caption{Algorithm for classifier-free guidance sampling}\label{sup_clsf_free_sample}
\hspace*{\algorithmicindent} \textbf{Input} The protein binding pocket $P$, learned diffusion model $\phi_{\theta}$, regression model for binding affinity prediction $f_{\psi}$, target binding affinity $\Delta G_{target}$, scale factor on guidance $s$ \\
\hspace*{\algorithmicindent} \textbf{Output} Sampled ligand molecule $M$ that binds to pocket $P$
\begin{algorithmic}
\State Sample number of atoms in $M$ based on the prior distribution conditioned on pocket size
\State Move the center of mass of protein pocket $P$ to zero, do the same movement for ligand $M$
\State Sample initial molecular atom coordinates $x_T$ and atom types $v_T$
\State $x_T \in \mathcal{N}(0,\textbf{I})$
\State $v_T = one\_hot(\arg\max_{i}(g_{i})), where\ g \sim Gumble(0,1)$
\For{t in $T,T-1,...,1$}
    \State Predict [$\hat{x}^c_0,\hat{v}_0$] with condition through [$\hat{x}_0,\hat{v}_0$] = $\phi_{\theta}$([$(x_t,\Delta G_{target}),v_t$], t, $P$)
    \State Predict [$\hat{x}_0,\hat{v}_0$] unconditionally through [$\hat{x}_0,\hat{v}_0$] = $\phi_{\theta}$([$(x_t,\varnothing),v_t$], t, $P$)
    \State Calculate guided term via: 
    
    $\hat{x}'_0 = (1-s)\hat{x}_0+s\hat{x}^c_0$  
    \State $\tilde{\mu}_t(x_t,\hat{x}'_0) = \frac{\sqrt{\Bar{\alpha}_{t-1}}\beta_t}{(1-\Bar{\alpha}_t)}\hat{x}'_0 + \frac{\sqrt{\alpha_t}(1-\Bar{\alpha}_{t-1})}{(1-\Bar{\alpha}_t)}x_t$
    \State $\Tilde{\beta}_t = \frac{1-\Bar{\alpha}_{t-1}}{1-\Bar{\alpha}_t}\beta_t$
    \State sample $\epsilon \sim \mathcal{N}(0,\textbf{I})$
    \State $x_{t-1} = \epsilon\sqrt{\Tilde{\beta}_t}+\tilde{\mu}_t(x_t,\hat{x}_0)$

    \State Sample $v_{t-1}$ from $q_{\theta}(v_{t-1}|v_t, \hat{v}_0) = \mathcal{C}(v_{t-1}|\Tilde{c}(v_t,v_0))$
    \State $\Tilde{c}(v_t,v_0)=(\alpha_t v_t+(1-\alpha_t)/K)\odot(\Bar{\alpha}_{t-1}v_0+(1-\Bar{\alpha}_{t-1})/K)$
    \State Sample $v_{t-1}$
    \Indent
    \State $v_{t-1} = \arg\max(\Tilde{c}(v_t,v_0))$
    \EndIndent
\EndFor
\end{algorithmic}
\end{algorithm}

\section{Algorithm for training diffusion model for classifier-free guidance}\label{alg:clsf_free_train_sup}

We outline the full algorithm for training our diffusion model for classifier free guidance, which is discussed in\mref 3.2.

\begin{algorithm}[H]
\caption{Algorithm for training diffusion model for classifier-free guidance}\label{clsf_free_train_alg}
\hspace*{\algorithmicindent} \textbf{Input} The protein-ligand binding dataset $\{(P_i,M_i),\Delta G_{i}\}^{N}_{i = 1}$, a neural network $\phi_{\theta}()$
\begin{algorithmic}
\While{$\phi_{\theta}()$ does not converge}
\State Sample diffusion time step $t \in \mathcal{U}(0,...,T)$
\State Move the complex to make CoM of protein atoms zero
\State Perturb $x_0$ to obtain $x_t$: $x_t = \sqrt{\Bar{\alpha}_t}x_0 + \sqrt{1-\Bar{\alpha}_t}\epsilon, $where $ \epsilon \in \mathcal{N}(0,I)$
\State Perturb $v_0$ to obtain $v_t$: 

$\log c = \log(\bar{\alpha}_t v_0 + (1-\bar{\alpha}_t )/K )$

$v_t = onehot(\arg\max_i[g_i + \log c_i]),$ where $ g \sim Gumbel(0, 1)$
\State compose data with binding affinity as conditioning $(x_t, \Delta G)$
\State random discard conditioning by $\Delta G = \varnothing$ with probability $p_{unconditional}$
\State Predict [$\hat{x}_0,\hat{v}_0$] through [$\hat{x}_0,\hat{v}_0$] = $\phi_{\theta}$([$(x_t, \Delta G),v_t$], t, $P$)
\State Compute the posterior atom types $c(v_t, v_0)$ and $c(v_t,  \hat{v}_0)$ according to Eq. 3 (main paper)
\State Compute the unweighted MSE loss on atom coordinates and the KL loss on posterior atom types: $\mathcal{L} = ||x_0 - \hat{x}_0||^2 +\alpha KL(c(v_t, v_0)||c(v_t, \hat{v}_0))$
\State Update $\theta$ by minimizing $\mathcal{L}$

\EndWhile

\end{algorithmic}
\end{algorithm}

\section{Implementation details}\label{impdet}

We provide further details on our implementation for the different components of our method. The classifier models are discussed in\mref 3.1.

\paragraph{Architecture Details}
For the architecture of the noise network and the binding energy prediction classifier, we both adopt the Uni-Mol architecture following \cite{zhou2023unimol_}. We also perform an ablation using EGNN~\cite{satorras2021n} as the classifier architecture. We outline the details for architecture in \tref{tab:model_architecture}.

\begin{table}[htbp]
\centering
\caption{Model Architecture Details}
\label{tab:model_architecture}
\begin{adjustbox}{width=\textwidth}
\begin{tabular}{lllll}
\toprule
\textbf{Component} & \textbf{Parameter} & \textbf{Uni-Mol Diffusion Model} & \textbf{EGNN Classifier} & \textbf{Uni-mol Classifier} \\
\midrule
\multirow{6}{*}{Network Architecture} 
& Node Feature Hidden Dimension & 128 & 128 & 128 \\
& Number of Layers & 9 & 2 & 9 \\
& Number of Attention Heads & 16 & 8 & 16 \\
& Edge Feature Dimension & 4 & 4 & 4 \\
& KNN number of nearest neighbors & 32 & 32 & 32 \\
\midrule
{Classifier Specific} 
& Pooling Method & - & Add & Mean \\
\bottomrule
\end{tabular}
\end{adjustbox}
\end{table}

\paragraph{Parameters for EGNN classifier model.}
The Equivariant Graph Neural Network (EGNN) is built based on~\citet{igashov2024equivariant}. The model contains two equivariant graph convolution layers. The total number of parameters for the model is $0.3$ million.

\paragraph{Training EGNN.}
The EGNN is trained using Adam~\cite{diederik2014adam}, with learning rate = $5e^{-4}$, weight decay = 0, $\beta_1$ = 0.95, and $\beta_2$=0.999. We use the ReduceLROnPlateau scheduler with decaying factor = 0.5, patience = 2 and minimum learning rate = $1e^{-6}$. We use a Mean Squared Error (MSE) loss. We train the model for 20 epochs, and the loss drops to $0.1$. 
For the loss, we apply loss masking to get rid of the invalid data. Specifically, for any data with a ground truth binding affinity $>0$ kcal/mol, we set the loss for this data to be zero during training. 

\paragraph{Training multi-constraints classifier model.}
The multi-properties regression model is mostly equivalent to the Binding Affinity Regression Model, with the primary difference being that the model has an output dimension of 3 for predicting Binding Affinity, Quantitative Drug Likeness (QED), and Synthetic
Accessibility (SA), respectively. In training, the ground truth and predicted binding affinities are both scaled by $-1/12$ to set them to approximately the same range as QED and SA. We assign equal weights to each property in the loss function equation. We use both Mean Absolute Error (MAE) and Mean Squared Error (MSE) loss to train the regression model, shown in Table~\ref{tab:multiprop-table}. We train the model for 60 epochs, with the loss approximately converged to $0.6$. We apply the same masking technique as in Binding Affinity Diffusion Guidance.
\paragraph{Parameters for Transformer classifier model.}
The Transformer is built based on~\citet{zhou2023unimol}. The model contains 10 attention layers. The total number of parameters for the model is $2.9$ million.

\paragraph{Training the Transformer.}
The Transformer is trained by using Adam~\cite{diederik2014adam}, with learning rate = $5e^{-4}$, weight decay = 0, $\beta_1$ = 0.95, and $\beta_2$=0.999. We use ReduceLROnPlateau scheduler with decaying factor = 0.5, patience = 2 and minimum learning rate = $1e^{-6}$. We use a Mean Squared Error (MSE) loss. We train the model for 20 epochs, and the loss drop down to $0.02$. 
For the loss, we apply loss masking to get rid of the invalid data. Specifically, for any data with a ground truth binding affinity $>0$ kcal/mol, we set the loss for this data to be zero during training. 

\paragraph{Parameters for the Diffusion model.}
For classifier guidance, we use the pre-trained checkpoint of the diffusion model from~\citet{guan20233d} and~\citet{guan2023decompdiff} for TargetDiff and DecompDiff, respectively. We apply our guidance method on top of these trained models.

For classifier-free guidance, we retrain the model using Algorithm~\ref{clsf_free_train_alg} with Transformer discussed in this section as architecture.

\paragraph{Diffusion sampling with guidance.}
During the sampling, we apply guidance with a certain combination of the scale factor and $\Delta G_{target}$. We apply clipping to the term $\frac{\beta_t}{\sqrt{\alpha_t}} w \nabla_{\boldsymbol{x}_{t}}\mathcal{L}(\Delta G_{predict}, \Delta G_{target})$ in~Eq.17 (main paper) to improve the stability of the sampling process. The hyperparameters for the results in \mtref 1 (\mref 5) are reported in~\tref{scc combination}.

Diffusion sampling takes 1000 steps. For "DecompDiff Ref + classifier guidance" and "DecompDiff Beta + classifier guidance" we report the metric for the results at sampled steps = 1000. For "TargetDiff + classifier guidance" we employ early stopping and report the results at sampled steps = 960.

\paragraph{Diffusion sampling with multi-constraints guidance}
In this part, we provide details on how the weight coefficients \( w_{\text{Vina}} \), \( w_{\text{QED}} \), and \( w_{\text{SA}} \) are chosen for the multi-constraint loss function. These weights are kept consistent between the classifier training stage and the inference (sampling) stage.

\begin{itemize}
    \item \textbf{Training stage loss weights.} We used uniform weights for all three properties, $w_{\text{Vina}} = w_{\text{QED}} = w_{\text{SA}} = 1$, with no grid search. Equal weights were chosen \emph{a priori} because binding affinity was linearly rescaled by $-1/12$, ensuring that its numerical range matches those of QED and SA during both training and inference (i.e., all lie approximately in $[0,1]$). Once all targets share a common scale, uniform weights provide a neutral multi-task objective that allows the regressor to learn each property equally, while keeping the setup simple, transparent, and reproducible.
    \item \textbf{Inference stage strategy.} At inference, we keep the same equal-weight objective for consistency with training. To emphasize different design goals, we adjust the conditioning/context targets---e.g., by specifying desired values for Vina, QED, and SA---rather than re-weighting the loss. This preserves the trained objective, avoids introducing additional hyperparameters, and yields a simple, effective, and reproducible control mechanism for navigating property trade-offs.
\end{itemize}

\begin{table}[h]
    \centering
    \caption{Scale factors and $\Delta G_{target}$ for the experiments reported in \mtref 1.}
    \begin{tabular}{c | c | c | c}
    \toprule
    Methods & Scale factor & $\Delta G_{target}$(kcal/mol) & Clipping\\
    \midrule
    TargetDiff + classifier guidance& 80 & -16 & 1 \\
    DecompDiff Ref + classifier guidance& 100 & -40 & 0.003 \\
    DecompDiff Beta + classifier guidance& 100 & -40 & 0.003 \\
    \bottomrule
    \end{tabular}
    \label{scc combination}
\end{table}

\paragraph{GPU information.}
All the experiments are conducted on an NVIDIA RTX 6000 Ada Generation.

\paragraph{Benchmark score calculations.}
We calculated QED, SA, and binding affinity using the same code base as in~\citet{guan20233d}. Diversity is calculated as follows for the sampled ligands, following~\citet{guan20233d,guan2023decompdiff}:
\begin{equation}
    \text{Diversity} = \frac{1}{n}\sum_{n}^{1}(1- \text{pairwise~Tanimoto~similarity}).
\end{equation}

\section{Other benchmark table}
We provide benchmarks of our method with \textbf{DecompOpt}. \tref{fair comparison} shows the benchmarking results with DecompOpt~\cite{zhou2023decompopt}. According to~\citet{zhou2023decompopt}, \textbf{DecompOpt} and \textbf{TargetDiff w/ Opt.} sample 600 ligands for each pocket and select the top 20 candidates filtered by AutoDock Vina. To compare with these approaches, we sample 100 ligands for each pocket and select the top 20 candidates to compute the final binding affinity performance. The results show that \modelnobf outperforms DecompOpt by up to 50\% in Vina Score, Vina Min, and Vina Dock.

\begin{table*}
\caption{We benchmark binding affinity performance with DecompOpt~\cite{zhou2023decompopt} on the same test set with 100 pockets. To compare with DecompOpt and TargetDiff w/ Opt. under the same conditions, we sample 100 ligands for each pocket. We then select the top 20 candidates to compute the final binding affinity performance.}
\label{fair comparison}
\centering
\resizebox{\textwidth}{!}{
\begin{tabular}{c|c|cc|cc}
\toprule
\multicolumn{2}{c|}{Method \textbar\ Metric} & \multicolumn{2}{c|}{Vina Score} & \multicolumn{2}{c}{Vina Min} \\
\multicolumn{2}{c|}{} & Mean~($\Delta\%$) & Med~($\Delta\%$) & Mean~($\Delta\%$) & Med~($\Delta\%$) \\
\midrule
\multirow{2}{*}{Diff.}
& TargetDiff & {-8.70} & -8.72 & -9.28 & -9.25 \\
& DecompDiff Beta~\cite{guan2023decompdiff} & -6.33 & -7.56 & -8.50 & -8.88 \\
\midrule
\multirow{2}{*}{Diff. + OPT.}
& \cellcolor{gray!25}TargetDiff w/ Opt.~\cite{zhou2023decompopt} & -7.87 & -7.48 & -7.82 & -7.48 \\
& \cellcolor{gray!25}DecompOpt~\cite{zhou2023decompopt} & -5.87 & -6.81 & -7.35 & -7.72 \\
\midrule
\multirow{4}{*}{Diff. + BADGER}
& TargetDiff + Classifier Guidance & \textbf{-10.51}~(+20.8\%) & \textbf{-11.12}~(+27.5\%) & \textbf{-10.99}~(+18.4\%) & \textbf{-11.22}~(+21.9\%) \\
& TargetDiff + Multi-Constraints Classifier Guidance & -8.86~(+1.8\%) & \underline{-9.28}~(+6.4\%) & {-9.48}~(+2.1\%) & {-9.67}~(+4.5\%) \\
& DecompDiff Beta + Classifier Guidance & -8.65~(+36.6\%) & {-9.68}~(+28.0\%) & \underline{-10.20}~(+20.0\%) & \underline{-10.49}~(+18.1\%) \\
& TargetDiff + Classifier-Free Guidance & \underline{-8.95}~(+2.8\%) & {-9.00}~(+3.2\%) & {-9.30}~(+0.2\%) & {-9.19}~(+0\%) \\
\bottomrule
\end{tabular}
}
\end{table*}

\section{Extra results on binding affinity improvement}\label{100pktappendix}
\begin{figure*}[htbp]
    \centering
    \includegraphics[width=0.98\linewidth]{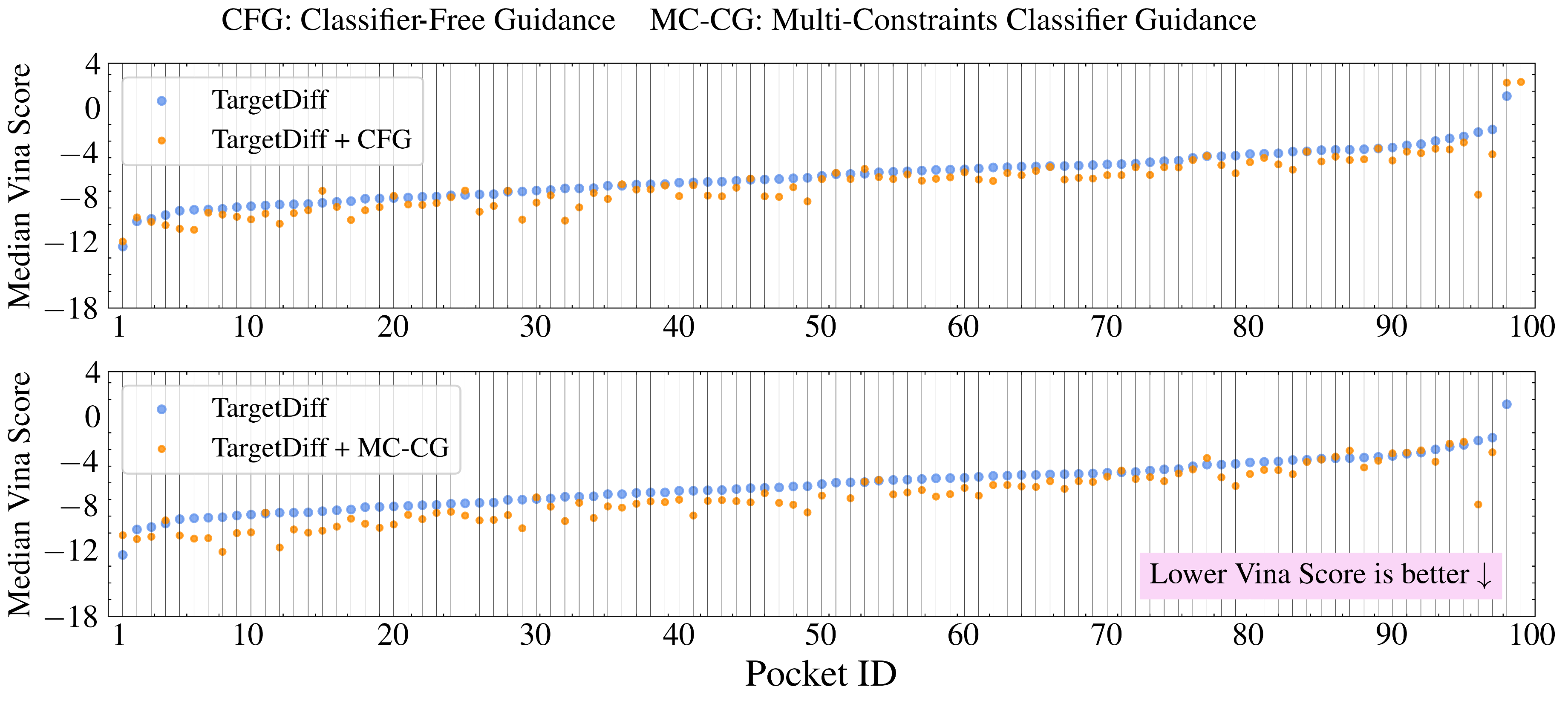}
  \caption{We visualize the improvement in median Vina Score on each of the 100 pockets in the test set for each diffusion model (TargetDiff, DecompDiff Ref, and DecompDiff Beta) after applying Classifier-Free Guidance (CFG) and Multi-Constraints Classifier Guidance (MC-CG) versions of  BADGER. BADGER improves the median Vina Score for most of the protein pockets. } 
  \label{pocketwise_extra}
\end{figure*}

We provide the improvement in median Vina Score results on Classifier-Free Guidance (CFG) and Multi-Constraints Classifier Guidance (MC-CG) in ~\fref{pocketwise_extra}.

\section{Extra results on Redocking RMSD for sampled molecules}

\begin{figure*}[htbp]
      \centering
      \includegraphics[width=\linewidth]{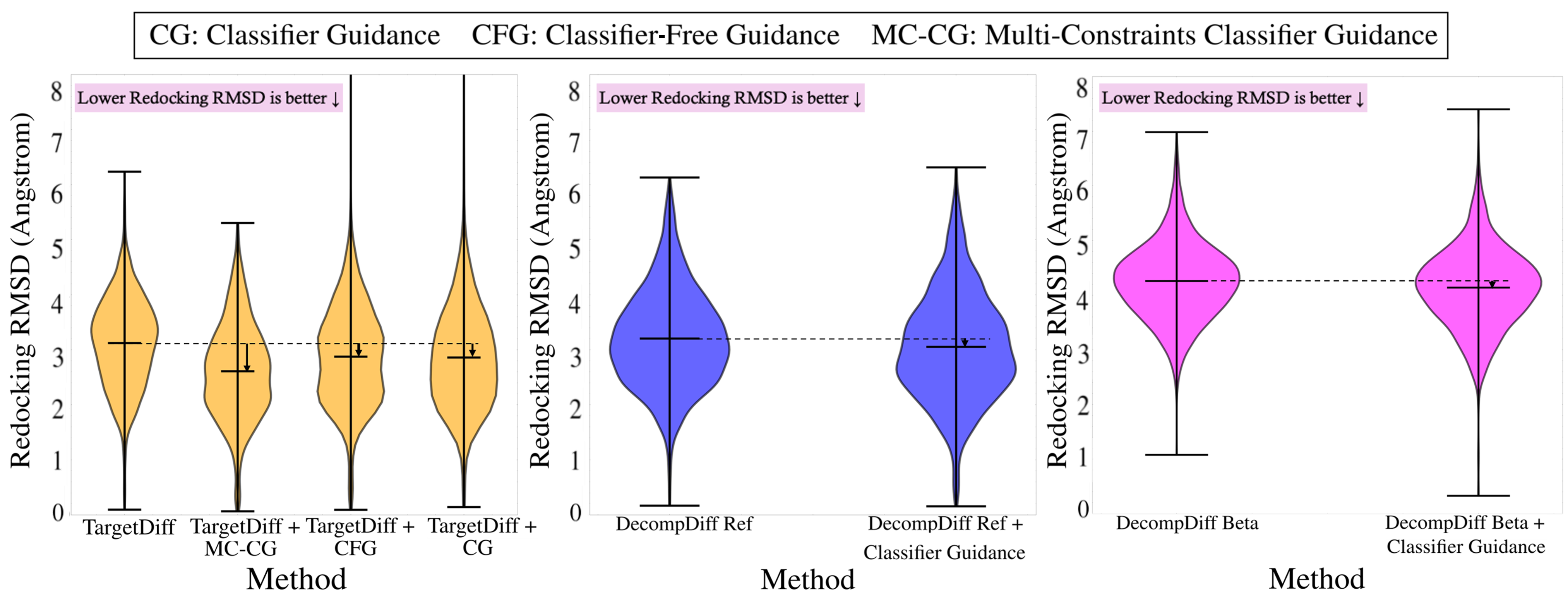}
      \label{rdrmsd1}

    \caption{\textbf{Redocking RMSD improvement with BADGER.} Redocking RMSD plot: lower redocking RMSD indicates that sampled poses have a better agreement with the Vina docking score function.}
    \label{redockrmsd}
\end{figure*}

\textit{Redocking RMSD} measures how closely the model-generated ligand pose matches the AutoDock Vina docked pose~\cite{harris2023benchmarking}. A lower redocking RMSD suggests better agreement between the pose before and after redocking, indicating that \modelnobf more accurately mimics the docking score function.~\fref{redockrmsd} compares redocking RMSD across models with and without \modelnobfns. The results show that \modelnobf lowers the RMSD, improving the quality of the ligand poses sampled from diffusion model.

\section{Consistency between vina score and experimental measured binding energy}
We compare the AutoDock Vina scores used for guiding molecular generation with experimentally measured binding affinities from PDBBind2016. We do not perform this comparison on CrossDock2020 because the dissociation constants in CrossDock2020 are derived from PDBBind, and not all entries contain experimentally measured values. In many cases, values are simply copied from PDBBind where there is an overlap in the protein pocket and ligand.

\begin{figure}
    \centering
    \includegraphics[width=0.7\linewidth]{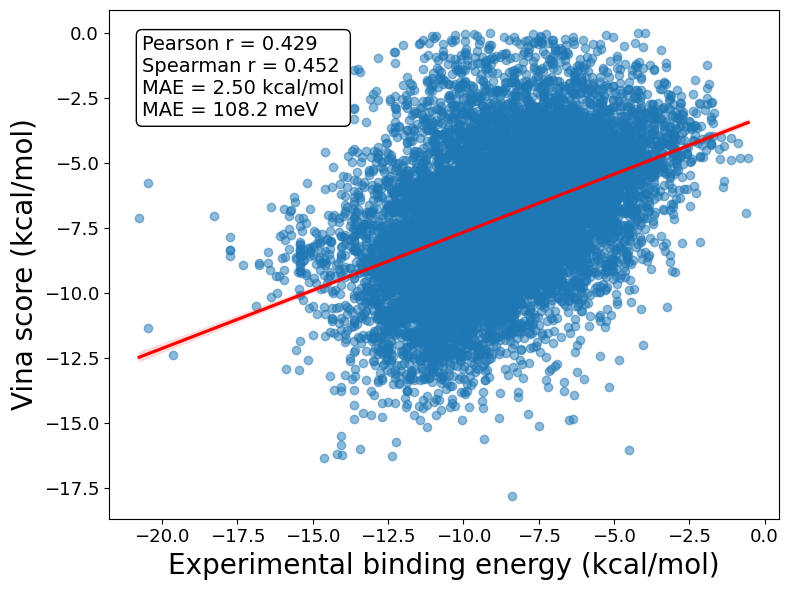}
    \caption{We analyze the consistency between Autodock Vina function and the experimental measured binding affinity value, specifically, we calculated the experimental binding energy using Ki/Kd/IC50 provided for each protein ligand complexes in PDBBind2016 using room temperature $T = 298.15K$. }
    \label{fig:pk-vs-vina}
\end{figure}

To perform the comparison, we convert the Ki/Kd/IC50 values in PDBBind2016 to $\Delta G$ using the equation:
\[
\Delta G = RT \ln K
\]
where $T = 298.15\,\text{K}$ (room temperature), and $R$ is the gas constant. The resulting $\Delta G$ values are then compared with the binding energies calculated using the AutoDock Vina score function, with both expressed in kcal/mol. As shown in \fref{fig:pk-vs-vina}, we observe a clear positive correlation between the Vina-calculated binding affinities and the experimentally measured values.

This result supports the use of the Vina score as a meaningful and computationally efficient proxy for binding affinity in our guidance framework. Nonetheless, we acknowledge that incorporating more accurate affinity predictors for guiding generation remains an important direction for future work, especially to better align with experimental binding free energies.

\section{Ablation of guidance strength \space$s$}
\begin{figure}
    \centering
    \includegraphics[width=0.65\linewidth]{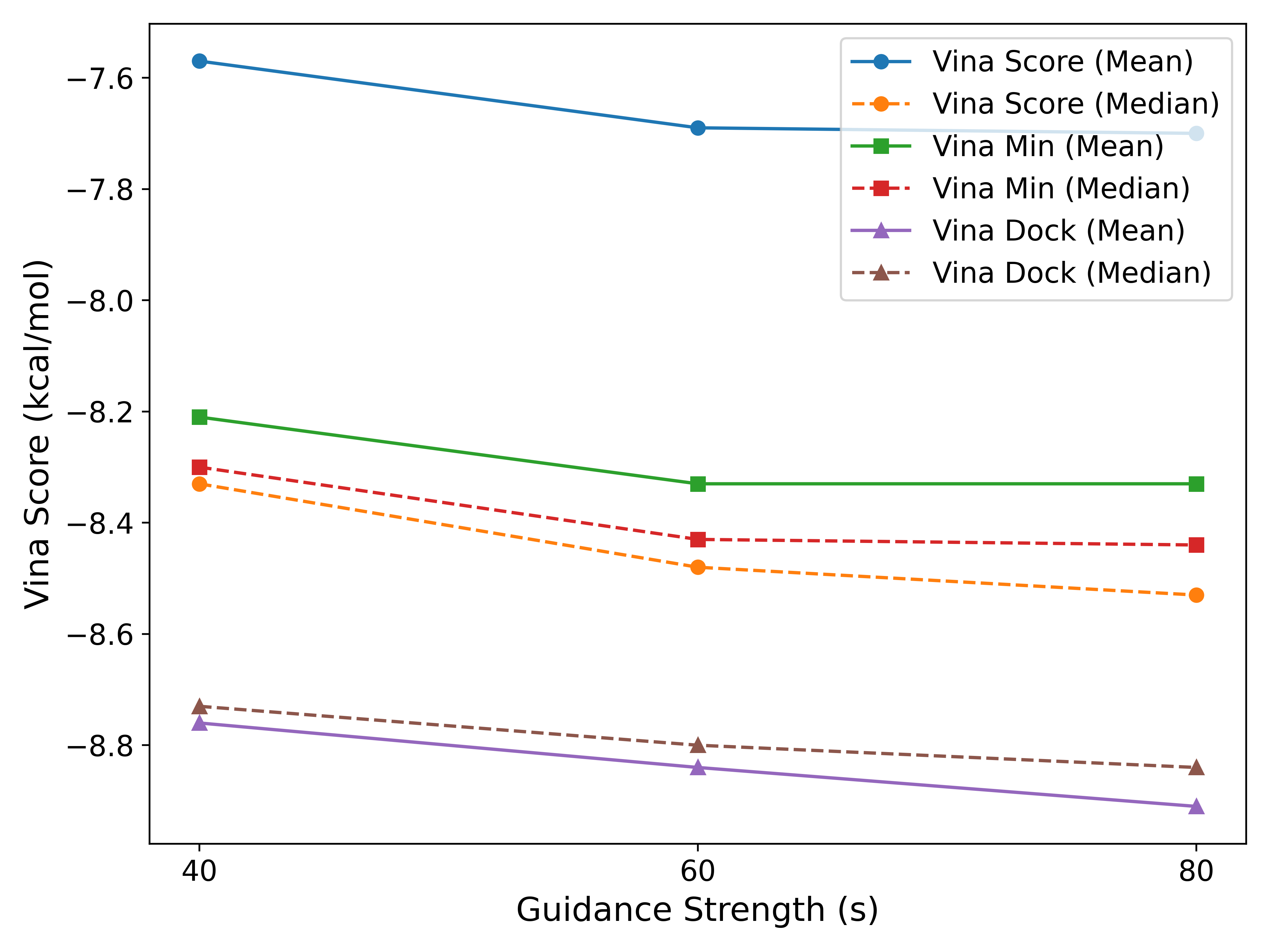}
    \caption{Vina scores (mean and median) under varying guidance strengths ($s \in \{40, 60, 80\}$) and fixed context value $c = -16$ for single-constraint guided TargetDiff sampling. As the guidance strength increases, it consistently result in lower (better) Vina scores across all three docking modes. However, the magnitude of improvement diminishes with higher $s$.}
    \label{fig::sampling-s-effect}
\end{figure}

To evaluate the impact of guidance strength $s$, we conducted an experiment where we sampled molecules using the same trained classifier and context value ($c = -16$), while varying $s \in \{40, 60, 80\}$. As shown in \tref{tab:sampling-guidance-sweep} and \fref{fig::sampling-s-effect}, increasing $s$ consistently led to lower (i.e., more favorable) Vina docking scores across all three docking modes (Score-Only, Minimize, Dock), demonstrating the effectiveness of stronger guidance. However, we also observed diminishing returns as $s$ increased, reflecting a balance between steering strength and adherence to the base generative distribution.

\begin{table}[ht]
    \centering
    \begin{tabular}{c|c|c|c}
        \textbf{Metric} & \textbf{s = 40} & \textbf{s = 60} & \textbf{s = 80} \\
        \hline
        Vina Score (Mean) & -7.57 & \underline{-7.69} & \textbf{-7.70} \\
        Vina Score (Median) & -8.33 & \underline{-8.48} & \textbf{-8.53} \\
        Vina Min (Mean) & -8.21 & \textbf{-8.33} & \textbf{-8.33} \\
        Vina Min (Median) & -8.30 & \underline{-8.43} & \textbf{-8.44} \\
        Vina Dock (Mean) & -8.76 & \underline{-8.84} & \textbf{-8.91} \\
        Vina Dock (Median) & -8.73 & \underline{-8.80} & \textbf{-8.84} \\
        QED (Mean) & \textbf{0.47} & \textbf{0.47} & \underline{0.46} \\
        QED (Median) & \textbf{0.48} & \underline{0.47} & 0.46 \\
        SA (Mean) & \textbf{0.51} & \underline{0.50} & \underline{0.50} \\
        SA (Median) & \textbf{0.50} & \underline{0.49} & \underline{0.49} \\
    \end{tabular}
    \caption{Sampling results for single-constraint-guided TargetDiff under varying guidance strengths ($s \in \{40, 60, 80\}$) and fixed context value $c = -16$. The best two results were highlighted with bold text and underlined text, respectively.}
    \label{tab:sampling-guidance-sweep}
\end{table}

\section{Benchmarking on Structural Validity, Specificity, Bond Angle, Bond Length and Chirality}

We benchmark the molecule structural validity for all baselines as well as our methods, the results are shown in \tref{tab:structural_validity}, to obtain the structural validity, we follow~\cite{schneuing2024structure} and calculate the percentage of the sampled molecules that pass the rdkit sanitize molecule function. The results shows that adding BADGER on top of other diffusion baseline doesn't affect the structural validity of the sampled molecules while also improved it's binding affinity.
\begin{table}[ht]
    \centering
    \begin{tabular}{c|c}
        \textbf{Model} & \textbf{Structural Validity}\% $\uparrow$ \\
        \hline
        AR & 92.95 \\
        liGAN & 99.11 \\
        Pocket2Mol & 98.31 \\
        TargetDiff & 98.96 \\
        DiffSBDD & 97.10 \\
        DecompDiff Ref & 75.48 \\
        DecompDiff Beta & 78.92 \\
        TargetDiff + Classifier Guidance & 97.37 \\
        TargetDiff + Multi-Constraints Classifier Guidance & 99.21 \\
        TargetDiff + Classifier-Free Guidance & 98.36 \\
        DecompDiff Ref + Classifier Guidance & 75.67 \\
        DecompDiff Beta + Classifier Guidance & 79.37 \\
    \end{tabular}
    \caption{Structural validity for each model, defined as percentage of sampled molecules that pass the rdkit sanitize function.}
    \label{tab:structural_validity}
\end{table}

We designed an evaluation procedure for specificity inspired by Gao et al. \cite{gao2024rethinking}, with modifications to improve robustness. Specifically, for each of the 100 protein pockets in the test set, we selected the top 10 generated molecules based on Vina docking scores. These molecules were then cross-docked into 5 randomly selected off-target pockets (with a consistent seed for all models), resulting in a total of 5{,}000 cross-docking evaluations. We discarded any docking results with invalid (positive) scores, and computed the average difference in docking energy between the on-target and off-target settings. This \textit{Specificity Score} quantifies specificity, with more negative values indicating greater difference hence greater selectivity for the intended target. As shown in \tref{tab:delta-score}, the classifier guidance framework significantly improves specificity, outperforming all baselines. We hypothesize that the addition of the classifier introduces richer structural conditioning signals, enabling the model to generate ligands that are more sensitive to the geometry and environment of the binding pocket. This observation is consistent with our earlier findings that guidance reduces steric clashes and improves redocking RMSD.
\begin{table}[ht]
    \centering
    \begin{tabular}{c|c}
        \textbf{Model} & \textbf{Specificity Score}$\downarrow$ \\
        \hline
        AR & -1.68 \\
        liGAN & -1.46 \\
        Pocket2Mol & -1.56 \\
        TargetDiff & -2.77 \\
        DiffSBDD & -1.89 \\
        DecompDiff Ref & -1.48 \\
        DecompDiff Beta & -2.82 \\
        TargetDiff + Classifier Guidance & \textbf{-4.28} \\
        TargetDiff + Multi-Constraints Classifier Guidance & -2.82 \\
        TargetDiff + Classifier-Free Guidance & -2.68 \\
        DecompDiff Ref + Classifier Guidance & -1.51 \\
        DecompDiff Beta + Classifier Guidance & \underline{-3.16} \\
    \end{tabular}
    \caption{Specificity Score for each model, defined as the mean difference in Vina docking score between the intended target and five off-target protein. More negative values indicate higher specificity of generated molecules for the intended target. The best two results were highlighted with bold text and underlined text, respectively.}
    \label{tab:delta-score}
\end{table}

We measured the Jensen-Shannon Divergence (JSD) between the bond length and bond angle distributions of generated molecules and those in the CrossDocked 2020 test set. Results are reported in \tref{tab:bond-angle-jsd} and \tref{tab:bond-length-jsd}. While some guided models show slightly worse alignment with reference distributions—particularly in the case of DecompDiff-guided variants—we also observe that certain configurations, most notably the multi-constraint guided TargetDiff model, yield improved JSD scores across multiple bond types.

These results suggest that our classifier-guided framework is capable of improving key target properties(eg. Vina score, QED, and synthetic accessibility), while still preserving, and in some cases enhancing, the chemical plausibility of generated molecules in terms of local geometric features. This further supports the applicability and usability of our approach in practical structure-based drug design settings.

\begin{table}[ht]
    \centering
    \resizebox{\textwidth}{!}{%
    \begin{tabular}{c|cccccc}
        \textbf{Model} & \textbf{CCC} & \textbf{CCO} & \textbf{CNC} & \textbf{NCC} & \textbf{CC=O} & \textbf{COC} \\
        \hline
        AR & 0.094 & 0.131 & 0.132 & 0.094 & 0.221 & 0.208 \\
        liGAN & 0.294 & 0.337 & 0.288 & 0.285 & 0.407 & 0.366 \\
        Pocket2Mol & 0.081 & 0.117 & \textbf{0.051} & 0.062 & \underline{0.123} & \textbf{0.100} \\
        TargetDiff & 0.072 & 0.091 & 0.078 & 0.056 & 0.124 & 0.150 \\
        DiffSBDD & 0.122 & 0.134 & 0.113 & 0.095 & 0.187 & 0.214 \\
        DecompDiff Ref & \textbf{0.053} & \underline{0.054} & 0.081 & \textbf{0.037} & 0.254 & 0.502 \\
        DecompDiff Beta & \underline{0.063} & \textbf{0.049} & 0.075 & \underline{0.044} & 0.368 & 0.509 \\
        TargetDiff + Classifier Guidance & 0.162 & 0.159 & 0.127 & 0.117 & 0.228 & 0.212 \\
        TargetDiff + Multi-Constraints Classifier Guidance & 0.079 & 0.094 & \underline{0.068} & 0.052 & \textbf{0.119} & \underline{0.140} \\
        TargetDiff + Classifier-Free Guidance & 0.180 & 0.164 & 0.199 & 0.172 & 0.250 & 0.304 \\
        DecompDiff Ref + Classifier Guidance & 0.080 & 0.072 & 0.145 & 0.062 & 0.296 & 0.501 \\
        DecompDiff Beta + Classifier Guidance & 0.089 & 0.080 & 0.133 & 0.069 & 0.317 & 0.444 \\
    \end{tabular}
    }
    \caption{Jensen-Shannon Divergence (JSD) between the bond angle distributions of generated molecules and the CrossDocked 2020 test set, across six common bond angle types. Lower JSD values indicate a closer match to the reference distribution. The best two results were highlighted with bold text and underlined text, respectively.}
    \label{tab:bond-angle-jsd}
\end{table}

\begin{table}[ht]
    \centering
    \resizebox{\textwidth}{!}{%
    \begin{tabular}{c|cccccccc}
        \textbf{Model} & \textbf{C-C} & \textbf{C=C} & \textbf{C:C} & \textbf{C-N} & \textbf{C=N} & \textbf{C:N} & \textbf{C-O} & \textbf{C=O} \\
        \hline
        AR & 0.568 & 0.425 & 0.454 & 0.386 & 0.437 & 0.483 & 0.394 & 0.507 \\
        liGAN & 0.580 & 0.488 & 0.508 & 0.593 & 0.641 & 0.594 & 0.599 & 0.626 \\
        Pocket2Mol & 0.436 & 0.304 & 0.427 & 0.324 & 0.381 & 0.414 & 0.328 & 0.459 \\
        TargetDiff & \textbf{0.298} & \underline{0.190} & 0.203 & \textbf{0.239} & \underline{0.168} & \textbf{0.128} & \textbf{0.299} & \underline{0.388} \\
        DiffSBDD & 0.368 & 0.313 & 0.337 & 0.318 & 0.332 & 0.282 & 0.355 & 0.400 \\
        DecompDiff Ref & 0.336 & 0.221 & \textbf{0.197} & 0.279 & 0.600 & 0.374 & 0.789 & 0.771 \\
        DecompDiff Beta & 0.397 & 0.259 & 0.229 & 0.256 & 0.603 & 0.414 & 0.779 & 0.729 \\
        TargetDiff + Classifier Guidance & 0.386 & 0.309 & 0.308 & 0.321 & 0.323 & 0.268 & 0.392 & \underline{0.388} \\
        TargetDiff + Multi-Constraints Classifier Guidance & \underline{0.304} & \textbf{0.182} & \underline{0.201} & \underline{0.245} & \textbf{0.159} & \underline{0.136} & \underline{0.300} & \textbf{0.377} \\
        TargetDiff + Classifier-Free Guidance & 0.438 & 0.368 & 0.393 & 0.374 & 0.431 & 0.368 & 0.423 & 0.440 \\
        DecompDiff Ref + Classifier Guidance & 0.443 & 0.327 & 0.339 & 0.326 & 0.618 & 0.414 & 0.735 & 0.722 \\
        DecompDiff Beta + Classifier Guidance & 0.484 & 0.354 & 0.381 & 0.305 & 0.616 & 0.501 & 0.729 & 0.713 \\
    \end{tabular}
    }
    \caption{Jensen-Shannon Divergence (JSD) between the bond length distributions of generated molecules and the CrossDocked 2020 test set, across eight bond types. “-”, “=”, and “:” represent single, double, and aromatic bonds, respectively. Lower JSD values indicate a closer match to the reference distribution. The best two results were highlighted with bold text and underlined text, respectively.}
    \label{tab:bond-length-jsd}
\end{table}

We benchmark Chirality through distributional evaluation and assess the Jensen-Shannon Divergence (JSD) between the number of chiral centers in generated molecules and those in the CrossDocked 2020 test set. This provides a proxy for how well the model captures the statistical characteristics of stereochemistry in realistic, drug-like compounds.

As shown in \tref{tab:chirality-jsd}, most models achieve a reasonably close match to the reference distribution. Notably, the DecompDiff Ref model and its guided variant perform best, with JSD values of 0.079 and 0.086, respectively. The multi-constraint guided TargetDiff model also improves upon its unguided counterpart. Which further highlights the ability of the framework to improve upon key metrics while maintaining reasonable stereochemical complexity.

\begin{table}[ht]
    \centering
    \begin{tabular}{c|c}
        \textbf{Model} & \textbf{JSD}$\downarrow$ \\
        \hline
        AR & 0.088 \\
        liGAN & 0.089 \\
        Pocket2Mol & 0.101 \\
        TargetDiff & 0.137 \\
        DiffSBDD & 0.147 \\
        DecompDiff Ref & \textbf{0.079} \\
        DecompDiff Beta & 0.226 \\
        TargetDiff + Classifier Guidance & 0.173 \\
        TargetDiff + Multi-Constraints Classifier Guidance & 0.118 \\
        TargetDiff + Classifier-Free Guidance & 0.108 \\
        DecompDiff Ref + Classifier Guidance & \underline{0.086} \\
        DecompDiff Beta + Classifier Guidance & 0.256 \\
    \end{tabular}
    \caption{Jensen-Shannon Divergence (JSD) between the distribution of the number of chiral centers in generated molecules and the CrossDocked 2020 test set. Lower JSD values indicate a closer match to the reference distribution. The best two results were highlighted with bold text and underlined text, respectively.}
    \label{tab:chirality-jsd}
\end{table}

\section{Ablation on guidance with imperfect classifier}

We conducted a robustness analysis using a classifier intentionally stopped early during training, with a high validation loss of approximately 4.4 kcal/mol on Vina score prediction. This “bad” classifier was then used for guidance under the same sampling configuration as the properly trained one ($s = 80$, $c = -16$). As shown in \tref{tab:bad-vs-good-classifier}, molecules generated with the poorly trained classifier exhibited significantly worse docking scores across all modes (Score-Only, Minimize, Dock). This confirms that binding energy prediction accuracy meaningfully affects sampling outcomes, as expected.

\begin{figure}
    \centering
    \includegraphics[width=0.65\linewidth]{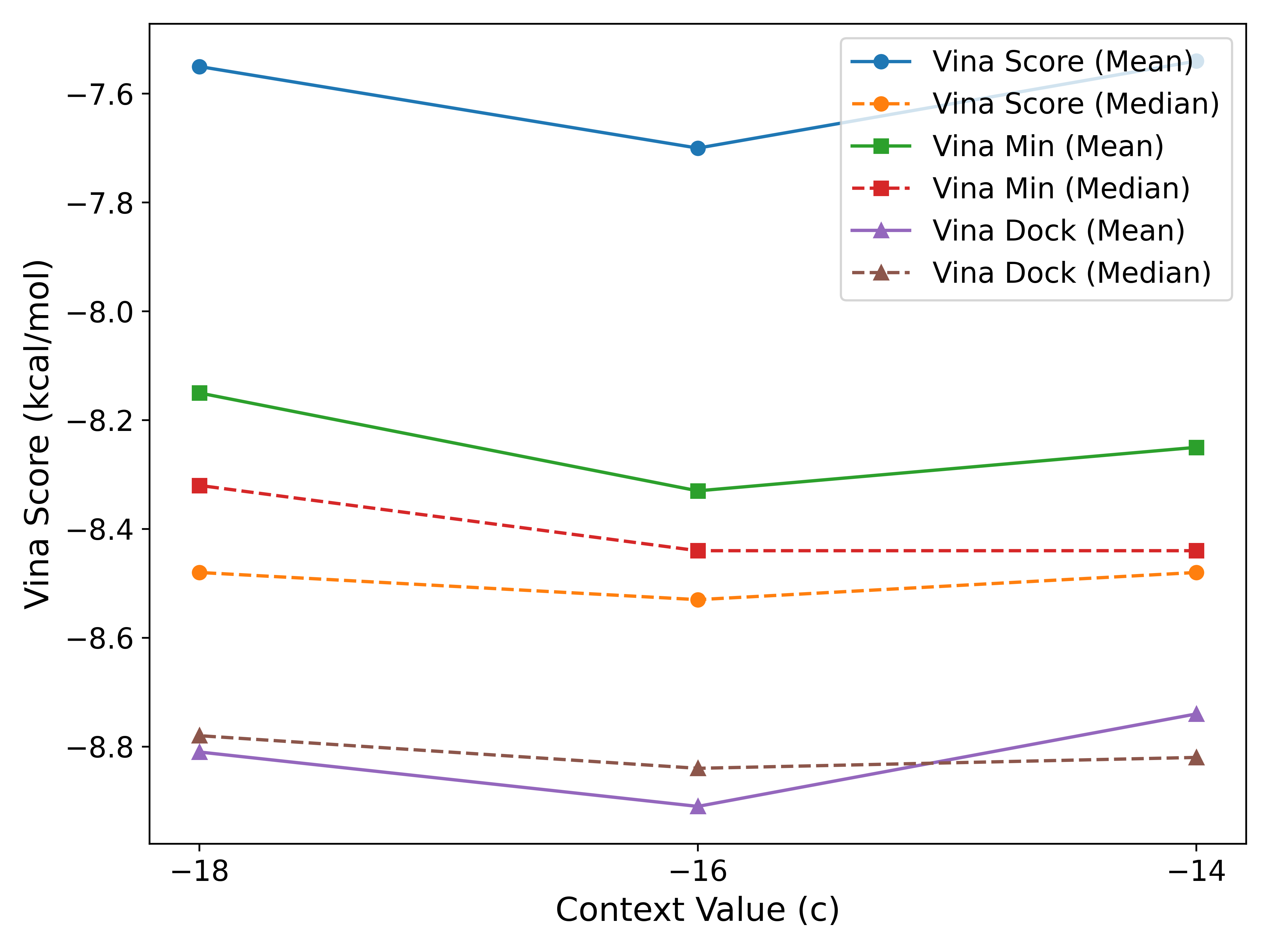}
    \caption{Vina scores (mean and median) under fixed guidance strength ($s = 80$) across different context values ($c = -14, -16, -18$) for single-constraint guided TargetDiff sampling. Scores improve from $c = -14$ to $c = -16$, but slightly worsen at $c = -18$.}
    \label{fig:revision-compare-c}
\end{figure}

In addition, we performed a sensitivity study on the context value $c$, which sets the optimization target for the classifier. Using a fixed guidance strength ($s = 80$), we varied $c \in \{-14, -16, -18\}$ to examine how aggressive guidance affect influence generation. As reported in \tref{tab:context-sweep} and \fref{fig:revision-compare-c}, performance improved as $c$ was lowered from $-14$ to $-16$, but declined slightly when pushed further to $-18$. These results suggest that while stronger guidance can enhance performance, overly aggressive or unreasonable classifier objectives can introduce instability or degrade quality.

\begin{table}[ht]
    \centering
    \begin{tabular}{c|c|c}
        \textbf{Metric} & \textbf{Good Classifier} & \textbf{Bad Classifier} \\
        \hline
        Vina Score (Mean) & -7.70 & -5.15 \\
        Vina Score (Median) & -8.53 & -5.90 \\
        Vina Min (Mean) & -8.33 & -6.40 \\
        Vina Min (Median) & -8.44 & -6.44 \\
        Vina Dock (Mean) & -8.91 & -7.59 \\
        Vina Dock (Median) & -8.84 & -7.62 \\
    \end{tabular}
    \caption{Sampling results for single-constraint guided TargetDiff using two classifiers with different training outcomes, both evaluated at guidance strength $s = 80$ and context $c = -16$. The “bad” classifier was intentionally stopped early at a high validation loss ($\sim$4.4 kcal/mol) on Vina score prediction. Poorly trained guidance significantly reduces docking performance across all Vina metrics.}
    \label{tab:bad-vs-good-classifier}
\end{table}

\begin{table}[ht]
    \centering
    \begin{tabular}{l|c|c|c}
        \textbf{Metric} & \textbf{c = -14} & \textbf{c = -16} & \textbf{c = -18} \\
        \hline
        Vina Score (Mean) & -7.54 & \textbf{-7.70} & \underline{-7.55} \\
        Vina Score (Median) & \underline{-8.48} & \textbf{-8.53} & \underline{-8.48} \\
        Vina Min (Mean) & \underline{-8.25} & \textbf{-8.33} & -8.15 \\
        Vina Min (Median) & \textbf{-8.44} & \textbf{-8.44} & \underline{-8.32} \\
        Vina Dock (Mean) & -8.74 & \textbf{-8.91} & \underline{-8.81} \\
        Vina Dock (Median) & \underline{-8.82} & \textbf{-8.84} & -8.78 \\
        QED (Mean) & \textbf{0.47} & \underline{0.46} & 0.44 \\
        QED (Median) & \textbf{0.47} & \underline{0.46} & 0.45 \\
        SA (Mean) & \textbf{0.50} & \textbf{0.50} & \textbf{0.50} \\
        SA (Median) & \textbf{0.50} & \underline{0.49} & \underline{0.49} \\
    \end{tabular}
    \caption{Sampling results for single-constraint guided TargetDiff under fixed guidance strength $s = 80$ and varying context values $c \in \{-14, -16, -18\}$. Increasing the context constraint (more negative $c$) initially improves docking scores, but extreme values (e.g., $c = -18$) begin to worsen scores. The best two results were highlighted with bold text and underlined text, respectively.}
    \label{tab:context-sweep}
\end{table}

\section{Benchmarking the speed of guidance}

To quantify the additional cost on sampling speed caused by guidance, we conducted benchmarking experiments under a consistent setup: all models were evaluated on an \textbf{NVIDIA Ada~6000} GPU with a batch size of \textbf{6} and \textbf{1000 diffusion sampling steps}. For fair comparison, the number of atoms in each sampled ligand was fixed to \textbf{60}. The parameter counts of each model are provided in \tref{tab:model_params}.  

The results in \tref{tab:speed} show that in our setting---where the diffusion model and classifier have comparable sizes---\textbf{classifier guidance} introduces a increase in computational cost relative to \textbf{classifier-free guidance}. This is expected, as classifier guidance requires an additional forward and backward pass through the classifier at every sampling step, whereas classifier-free guidance reuses the same diffusion network through two conditional forward passes per step. Nonetheless, the added cost remains moderate relative to the overall inference time.

\begin{table}[ht]
\centering
\begin{tabular}{l|c}
\toprule
\textbf{Model} & \textbf{Number of Parameters (M)} \\
\midrule
TargetDiff &  2.84 \\
TargetDiff Classifier&  2.89 \\
DecompDiff &  5.00 \\
DecompDiff Classifier&  2.89 \\
\bottomrule
\end{tabular}
\caption{Model parameter comparison.}
\label{tab:model_params}
\end{table}

\begin{table}[ht]
    \centering
    \begin{tabular}{c|c}
        \textbf{Model} & \textbf{Second per Step} $\downarrow$ \\
        \hline
        TargetDiff &  0.067\\
        DecompDiff &  0.173\\
        TargetDiff + Classifier Guidance &  0.201 \\
        TargetDiff + Classifier-Free Guidance &  0.149\\
        DecompDiff + Classifier Guidance &  0.575\\
        DecompDiff + Classifier-Free Guidance &  0.341\\
    \end{tabular}
    \caption{We evaluate the efficiency of different diffusion models and their guided variants by measuring the \textbf{average sampling time per diffusion sampling step} (in seconds). The benchmark reflects the computational overhead introduced by classifier or classifier-free guidance during generation. Lower values indicate faster sampling speed.}
    \label{tab:speed}
\end{table}

\section{Analysis of Poor Binding in Outlier Pockets}

In Figure~2 of the main paper, we observe that for a small number of binding pockets, unguided models produce consistently high (i.e., poor) Vina scores. Although BADGER substantially improves performance on these cases, these pockets remain challenging overall. In this appendix section, we investigate potential factors contributing to this behavior.

To systematically analyze these challenging cases, we identify a set of 20 outlier pockets from the test set. A pocket is classified as an outlier if its median Vina score exceeds $-1$\,kcal/mol for at least one of the following unguided baselines:
\textit{TargetDiff}, \textit{TargetDiff+CG}, \textit{TargetDiff+CFG}, \textit{TargetDiff+Multi-Constraint CG}, \textit{DecompDiff Ref}, \textit{DecompDiff Ref+CG}, \textit{DecompDiff Beta}, and \textit{DecompDiff Beta+CG}.

We then analyzed three structural properties commonly associated with difficult binding environments: pocket size, pocket hydrophobicity, and pocket depth. These were computed as follows:
\begin{itemize}
    \item \textbf{Pocket size (radius):} minimum distance between the ligand center and any heavy atom in the protein.
    \item \textbf{Pocket hydrophobicity:} fraction of hydrophobic residues within 5~\AA\ of the ligand center, using the set \{ALA, VAL, LEU, ILE, MET, PHE, TRP, TYR, PRO\}.
    \item \textbf{Pocket depth:} for residues within radius $r$ of the ligand center, we estimate depth as
    \[
    \max\bigl(\mathrm{dist}(\text{protein center}, \text{pocket heavy atoms})\bigr)
    \;-\;
    \mathrm{dist}(\text{protein center}, \text{ligand center}).
    \]
\end{itemize}

\begin{figure}[h]
    \centering
    \includegraphics[width=0.9\linewidth]{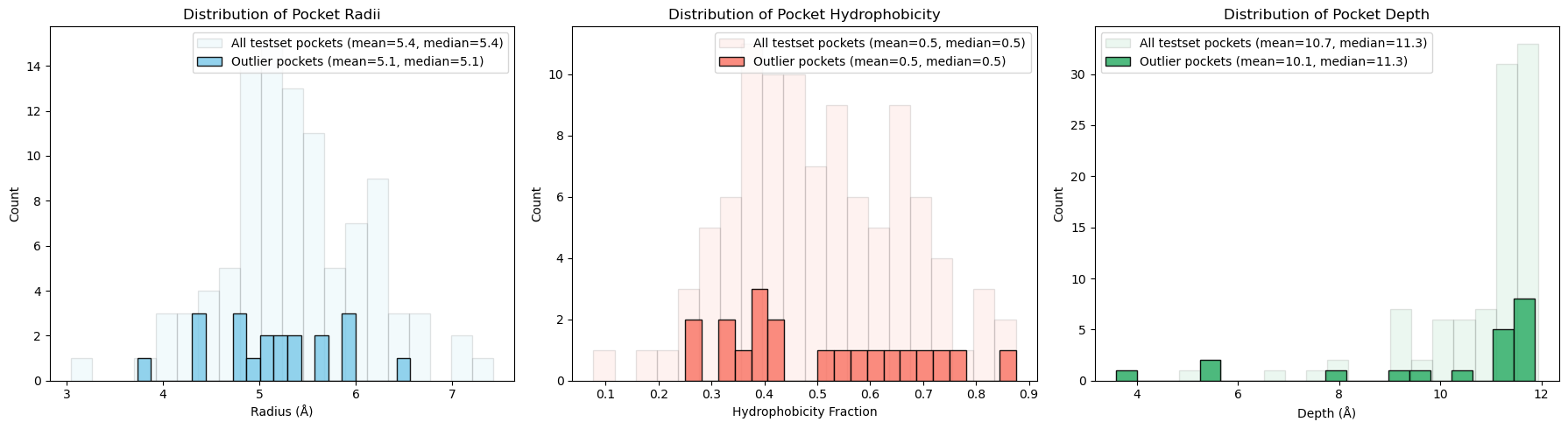}
    \caption{Distribution of pocket radii, hydrophobicity, and depth for all test-set pockets versus the outlier subset.}
    \label{pocket_analysis}
\end{figure}

Figure~\ref{pocket_analysis} compares the distributions for the full test set and the outlier subset. The outlier pockets tend to be \textbf{smaller and shallower}, indicating that diffusion models have more difficulty generating high-affinity poses when the pocket offers limited geometric enclosure. Hydrophobicity differences appear modest.

Finally, we note that pocket properties are not the only possible source of these outliers. In our experiments, we directly used pretrained diffusion-model checkpoints from prior work for both TargetDiff and DecompDiff. It is therefore possible that training factors (e.g., number of epochs, learning-rate schedule, optimizer choice) also limit model generalization on these more challenging pockets.

\section{Potential Reasons for Improved Diffusion Guidance with Clean-Data Classifier Training}
As discussed in the Methods section of the main paper, we observe that training the binding-affinity classifier on ``clean'' data can lead to improved guidance performance. In this appendix section, we further discuss potential reasons for this behavior. Specifically, we organize the discussion into two parts, focusing on effects arising during the classifier training stage and during the sampling stage.

Let $x_t$ denote the noised data produced by the forward diffusion process, where $t = 0$ corresponds to clean data and $t = T$ corresponds to pure Gaussian noise.

\paragraph{Training stage.}  
Classifier quality directly affects the accuracy of the $P(y \mid x_t)$ term used during guidance. Training the classifier on $x_0$ provides two benefits:
\begin{itemize}
    \item The classifier converges more easily, because clean inputs make it easier to distinguish different ligands, leading to a better fit to the dataset.
    \item The classifier performs better on low-noise samples $x_t$ with $t$ close to 0, which are the most critical steps during guidance.
\end{itemize}
In contrast, training on noisy inputs $x_t$ makes learning more difficult: as $t$ approaches $T$, ligands become nearly indistinguishable from Gaussian noise, making the regression task significantly harder.

\paragraph{Sampling stage.}  
During sampling, we apply gradient clipping to the classifier’s gradients, which prevents large, unstable gradients at high timesteps from corrupting generation. At low timesteps, where guidance is most influential, a classifier trained on clean $x_0$ performs better than one trained on noisy $x_t$. This leads to the empirical observation that training on $x_0$ improves guidance quality compared to training on noisier states.

\paragraph{Further discussion.}  
We note that our largest dataset contains 100{,}000 examples. The difference in training difficulty between $x_0$ and $x_t$ may diminish with substantially larger datasets or higher-capacity models. The relative performance of classifiers trained on $x_0$ vs.\ $x_t$ remains an interesting direction for future investigation.

\FloatBarrier
\newpage
\bibliographystyle{unsrtnat}
\bibliography{citation}
\end{document}